\documentclass[10pt,twocolumn,letterpaper]{article}

\usepackage{cvpr}
\usepackage{times}
\usepackage{epsfig}
\usepackage{graphicx}
\usepackage{amsmath}
\usepackage{amssymb}
\usepackage{float}
\usepackage{subfig}
\usepackage{url}
\usepackage{adjustbox}
\usepackage{color}

\usepackage{amsmath}
\usepackage{amssymb}
\usepackage{amsthm}

\newcommand{\calA}{\mathcal{A}}
\newcommand{\calI}{\mathcal{I}}
\newcommand{\calX}{\mathcal{X}}

\newcommand{\eat}[1]{}
\newcommand{\todo}[1]{\textcolor{red}{#1}}

\usepackage[pagebackref=false,breaklinks=true,colorlinks,bookmarks=false]{hyperref}

\cvprfinalcopy 

\def\cvprPaperID{2022} 

\begin{document}

\title{Towards Accurate Multi-person Pose Estimation in the Wild}

\author{%
George~Papandreou,
Tyler~Zhu,
Nori~Kanazawa,
Alexander~Toshev,\\
Jonathan~Tompson,
Chris~Bregler,
Kevin~Murphy\\
Google, Inc.\\
{\tt\small [gpapan, tylerzhu, kanazawa, toshev, tompson, bregler, kpmurphy]@google.com}
}

\maketitle

\begin{abstract}
We propose a method for multi-person detection and 2-D pose estimation that achieves state-of-art results on the challenging COCO keypoints task. It is a simple, yet powerful, top-down approach consisting of two stages.

In the first stage, we predict the location and scale of boxes which are likely to contain people; for this we use the Faster RCNN detector. In the second stage, we estimate the keypoints of the person potentially contained in each proposed bounding box. For each keypoint type we predict dense heatmaps and offsets using a fully convolutional ResNet. To combine these outputs we introduce a novel aggregation procedure to obtain highly localized keypoint predictions. We also use a novel form of keypoint-based Non-Maximum-Suppression (NMS), instead of the cruder box-level NMS, and a novel form of keypoint-based confidence score estimation, instead of box-level scoring.

Trained on COCO data alone, our final system achieves average precision of $0.649$ on the COCO test-dev set and the $0.643$ test-standard sets, outperforming the winner of the 2016 COCO keypoints challenge and other recent state-of-art. Further, by using additional in-house labeled data we obtain an even higher average precision of $0.685$ on the test-dev set and $0.673$ on the test-standard set, more than $5\%$ absolute improvement compared to the previous best performing method on the same dataset.
\end{abstract}

\section{Introduction}

Visual interpretation of people plays a central role in the quest for comprehensive image understanding. We want to localize people, understand the activities they are involved in, understand how people move for the purpose of Virtual/Augmented Reality, and learn from them to teach autonomous systems. A major cornerstone in achieving these goals is the problem of human pose estimation, defined as 2-D localization of human joints on the arms, legs, and keypoints on torso and the face.

Recently, there has been significant progress on this problem, mostly by leveraging deep Convolutional Neural Networks (CNNs) trained on large labeled datasets~\cite{deeppose, jainiclr2014, tompsonnips2014, Chen_NIPS14, stackedhourglass, andriluka14cvpr, bulat2016, zisserman2016, chain16, deeper_cut, cmu_mscoco}. However, most prior work has focused on the simpler setting of predicting the pose of a single person assuming the location and scale of the person is provided in the form of a ground truth bounding box or torso keypoint position, as in the popular MPII~\cite{andriluka14cvpr} and FLIC~\cite{modec2013} datasets.

In this paper, we tackle the more challenging setting of pose detection `in the wild', in which we are not provided with the ground truth location or scale of the person instances. This is harder because it combines the problem of person detection with the problem of pose estimation. In crowded scenes, where people are close to each other, it can be quite difficult to solve the association problem of determining which body part belongs to which person.

The recently released COCO person keypoints detection dataset and associated challenge \cite{keypointchallenge} provide an excellent vehicle to encourage research, establish metrics, and measure progress on this task. It extends the COCO dataset \cite{lin2014microsoft} with additional annotations of 17 keypoints (12 body joints and 5 face landmarks) for every medium and large sized person in each image. A large number of persons in the dataset are only partially visible. The degree of match between ground truth and predicted poses in the COCO keypoints task is measured in terms of object keypoint similarity (OKS), which ranges from 0 (poor match) to 1 (perfect match). The overall quality of the combined person detection and pose estimation system in the benchmark is measured in terms of an OKS-induced average precision (AP) metric. In this paper, we describe a system that achieves state-of-the-art results on this challenging task.

There are two broad approaches for tackling the multi-person pose estimation problem: \emph{bottom-up}, in which keypoint proposals are grouped together into person instances, and \emph{top-down}, in which a pose estimator is applied to the output of a bounding-box person detector. Recent work \cite{deepcut, deeper_cut, cmu_mscoco, insafutdinov2016articulated} has advocated the bottom-up approach; in their experiments, their proposed bottom-up methods outperformed the top-down baselines they compared with.

In contrast, in this work we revisit the top-down approach and show that it can be surprisingly effective. The proposed system is a two stage pipeline with state-of-art constituent components carefully adapted to our task. In the first stage, we predict the location and scale of boxes which are likely to contain people. For this we use the Faster-RCNN method \cite{Ren2015} on top of a ResNet-101 CNN \cite{He2016ResNets}, as implemented by \cite{huang2016speed}. In the second stage, we predict the locations of each keypoint for each of the proposed person boxes. For this we use a ResNet~\cite{He2016ResNets} applied in a fully convolutional fashion to predict activation heatmaps and offsets for each keypoint, similar to the works of Pishchulin et al.~\cite{deepcut} and Insafutdinov et al.~\cite{deeper_cut}, followed by combining their predictions using a novel form of heatmap-offset aggregation. We avoid duplicate pose detections by means of a novel keypoint-based Non-Maximum-Suppression (NMS) mechanism building directly on the OKS metric (which we call OKS-NMS), instead of the cruder box-level IOU NMS. We also propose a novel keypoint-based confidence score estimator, which we show leads to greatly improved AP compared to using the Faster-RCNN box scores for ranking our final pose proposals. The system described in this paper is an improved version of our G-RMI entry to the COCO 2016 keypoints detection challenge.

Using only publicly available data for training, our final system achieves average precision of $0.649$ on the COCO \emph{test-dev} set and $0.643$ on the COCO \emph{test-standard} set, outperforming the winner of the 2016 COCO keypoints challenge \cite{cmu_mscoco}, which gets $0.618$ on \emph{test-dev} and $0.611$ on \emph{test-standard}, as well as the very recent Mask-RCNN \cite{he2017mask} methods which gets $0.631$ on \emph{test-dev}. Using additional in-house labeled data we obtain an even higher average precision of $0.685$ on the test-dev set and $0.673$ on the test-standard set, more than $5\%$ absolute performance improvement over the best previous methods. These results have been attained with single-scale evaluation and using a single CNN for box detection and a single CNN for pose estimation. Multi-scale evaluation and CNN model ensembling might give additional gains.

In the rest of the paper, we discuss related work and then describe our method in more detail. We then perform an experimental study, comparing our system to recent state-of-the-art, and we measure the effects of the different parts of our system on the AP metric.



\section{Related Work}


For most of its history, the research in human pose estimation has been heavily based on the idea of part-based models, as pioneered by the Pictorial Structures (PS) model of Fischler and Elschlager~\cite{Fischler73}. One of the first practical and well performing methods based on this idea is Deformable Part Model (DPM) by Felzenswalb et al.~\cite{FelzenszwalbDPM}, which spurred a large body of work on probabilistic graphical models for 2-D human pose inference~\cite{andriluka2009pictorial, BetterAppearancePic, Sapp2010, yang11cvpr, dantone13cvpr, johnson11cvpr, pishchulin13cvpr, modec2013, armlets2013}. The majority of these methods focus on developing tractable inference procedures for highly articulated models, while at the same time capturing rich dependencies among body parts and properties.

\textbf{Single-Person Pose}
With the development of Deep Convolutional Neural Networks (CNN) for vision tasks, state-of-art performance on pose estimation is achieved using CNNs~\cite{deeppose, jainiclr2014, tompsonnips2014, Chen_NIPS14, stackedhourglass, andriluka14cvpr, bulat2016, zisserman2016, chain16, deeper_cut, cmu_mscoco}.
The problem can be formulated as a regression task, as done by Toshev and Szegedy~\cite{deeppose}, using a cascade of detectors for top-down pose refinement from cropped input patches. 
Alternatively, Jain et al.~\cite{jainiclr2014} trained a CNN on image patches, which was applied convolutionally at inference time to infer heatmaps (or activity-maps) for each keypoint independently. In addition, they used a ``DPM-like'' graphical-model post processing step to filter heatmap potentials and to impose inter-joint consistency. Following this work, Tompson et al.~\cite{tompsonnips2014} used a multi-scale fully-convolutional architecture trained on whole images (rather than image crops) to infer the heatmap potentials, and they reformulated the graphical model from \cite{jainiclr2014} - simplifying the tree structure to a star-graph and re-writing the belief propagation messages - so that the entire system could be trained end-to-end.

Chen et al.~\cite{Chen_NIPS14} added image-dependent priors to improve CNN performance. By learning a lower-dimensional image representation, they clustered the input image into a mixture of configurations of each pair of consecutive joints. Depending on which mixture is active for a given input image, a separate pairwise displacement prior was used for graphical model inference, resulting in stronger pairwise priors and improved overall performance. 

Bulat et al.~\cite{bulat2016} use a cascaded network to explicitly infer part relationships to improve inter-joint consistency, which the authors claim effectively encodes part constraints and inter-joint context. Similarly, Belagiannis \& Zisserman~\cite{zisserman2016} also propose a cascaded architecture to infer pairwise joint (or part) locations, which is then used to iteratively refine unary joint predictions, where unlike \cite{bulat2016}, they propose iterative refinement using a recursive neural network.

Inspired by recent work in sequence-to-sequence modeling, Gkioxari et al.~\cite{chain16} propose a novel network structure where body part locations are predicted sequentially rather than independently, as per traditional feed-forward networks. Body part locations are conditioned on the input image and all other predicted parts, yielding a model which promotes sequential reasoning and learns complex inter-joint relationships.

The state-of-the-art approach for single-person pose on the MPII human pose~\cite{andriluka14cvpr} and FLIC~\cite{modec2013} datasets is the CNN model of Newell et al.~\cite{stackedhourglass}. They propose a novel CNN architecture that uses skip-connections to promote multi-scale feature learning, as well as a repeated pooling-upsampling (``hourglass'') structure that results in improved iterative pose refinement. They claim that their network is able to more efficiently learn various spatial relationship associated with the body, even over large pixel displacements, and with a small number of total network parameters.

\textbf{Top-Down Multi-Person Pose}
The problem of multi-person pose estimation presents different challenges, unadressed by the above work. Most of the approaches for multi-person pose aim at associating person part detections with person instances. The \textit{top down} way to establish these associations, which is closest to our approach, is to first perform person detection followed by pose estimation. For example, Pishchulin et al.~\cite{pishchulin2012articulated} follow this paradigm by using PS-based pose estimation. A more robust to occlusions person detector, modeled after poselets, is used by Gkioxari et al.~\cite{gkioxari2014using}. Further, Yang and Ramanan~\cite{yang11cvpr} fuse detection and pose in one model by using a PS model. The inference procedure allows for pose estimation of multiple person instances per image analogous to PS-based object detection. A similar multi-person PS with additional explicit occlusion modeling is proposed by Eichner and Ferrari~\cite{eichner2010we}. The very recent Mask-RCNN method~\cite{he2017mask} extends Faster-RCNN~\cite{Ren2015} to also support keypoint estimation, obtaining very competitive results. On a related note, 2-D person detection is used as a first step in several 3D pose estimation works~\cite{sun2011articulated,belagiannis20143d,belagiannis20153d}. 

\textbf{Bottom-Up Multi-Person Pose}
A different line of work is to detect body parts instead of full persons, and to subsequently associate these parts to human instances, thus performing pose estimation in a \textit{bottom up} fashion. Such approaches employ part detectors and differ in how associations among parts are expressed, and the inference procedure used to obtain full part groupings into person instances. Pishchulin et al.~\cite{deepcut} and later Insafutdinov et al.~\cite{deeper_cut, insafutdinov2016articulated} formulate the problem of pose estimation as part grouping and labeling via a Linear Program. A similar formulation is proposed by Iqbal et al.~\cite{iqbal2016multi}. A probabilistic approach to part grouping and labeling is also proposed by Ladicky et al.~\cite{ladicky2013human}, leveraging a HOG-based system for part detections. 

Cao et al.~\cite{cmu_mscoco} winning entry to the 2016 COCO person keypoints challenge~\cite{lin2014microsoft} combines a variation of the unary joint detector architecture from \cite{wei2016convolutional} with a part affinity field regression to enforce inter-joint consistency. They employ a greedy algorithm to generate person instance proposals in a bottom-up fashion. Their best results are obtained in an additional top-down refinement process in which they run a standard single-person pose estimator \cite{wei2016convolutional} on the person instance box proposals generated by the bottom-up stage.


\section{Methods}
\label{cascaded_pose_estimation_model}

\begin{figure}
\centering
\includegraphics[width=0.45\textwidth,trim={1cm 0 0 0},clip]{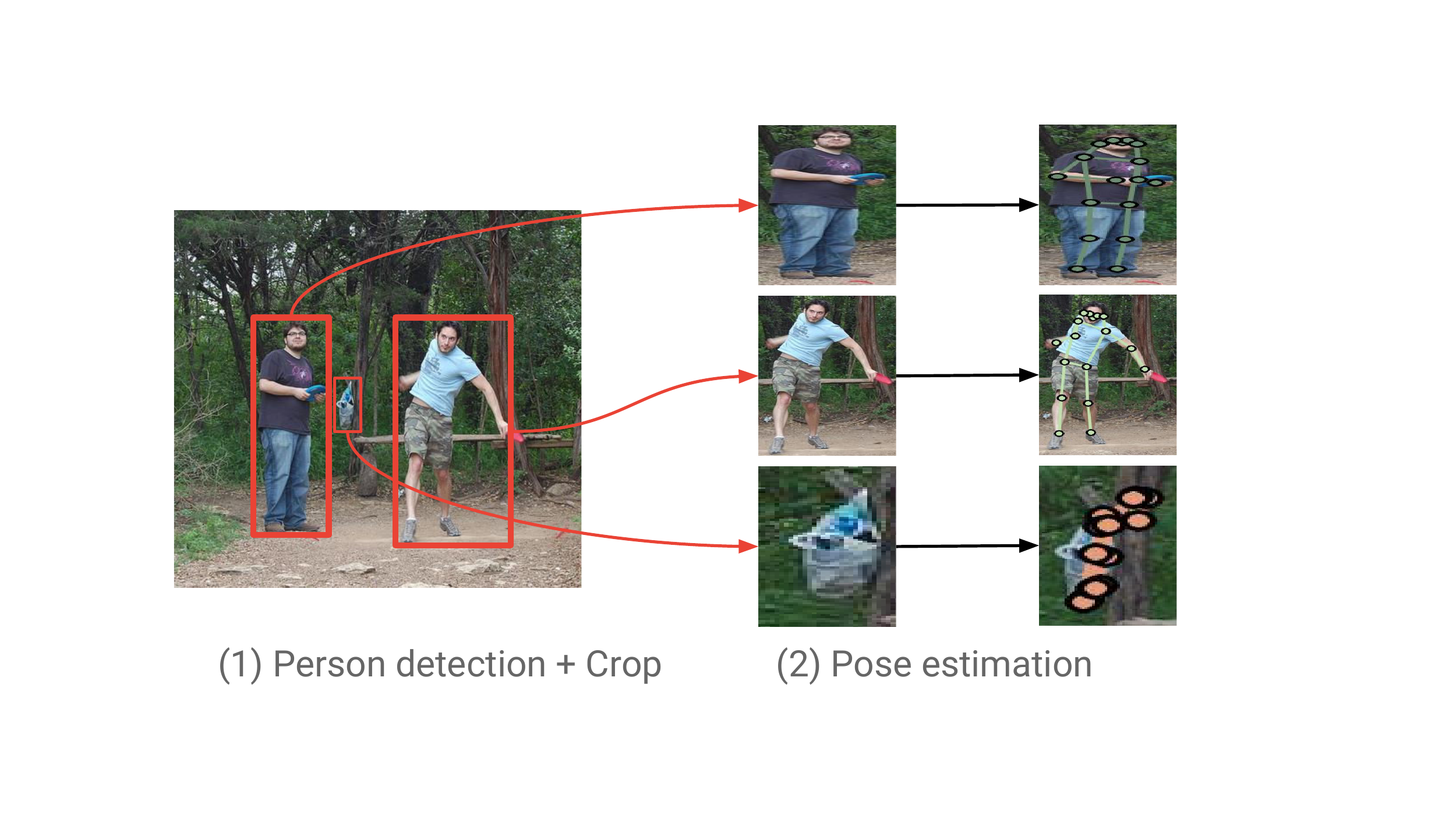}
\caption{\label{fig:grmi-overview}Overview of our two stage cascade model. In the first stage, we employ a Faster-RCNN person detector to produce a bounding box around each candidate person instance. In the second stage, we apply a pose estimator to the image crop extracted around each candidate person instance in order to localize its keypoints and re-score the corresponding proposal.}
\end{figure}

Our multi-person pose estimation system is a two step cascade, as illustrated in Figure~\ref{fig:grmi-overview}.

Our approach is inspired by recent state-of-art object detection systems such as \cite{girshick2014rich, szegedy2015going}, which propose objects in a class agnostic fashion as a first stage and refine their label and location in a second stage. We can think of the first stage of our method as a proposal mechanism, however of only one type of object -- person. Our second stage serves as a refinement where we (i) go beyond bounding boxes and predict keypoints and (ii) rescore the detection based on the estimated keypoints. For computational efficiency, we only forward to the second stage person box detection proposals with score higher than 0.3, resulting in only 3.5 proposals per image on average. In the following, we describe in more detail the two stages of our system.

\subsection{Person Box Detection}
\label{person_detection_model}

Our person detector is a Faster-RCNN system \cite{Ren2015}. In all experiments reported in this paper we use a ResNet-101 network backbone~\cite{He2016ResNets}, modified by atrous convolution \cite{chen2016deeplab, li2016r} to generate denser feature maps with output stride equal to 8 pixels instead of the default 32 pixels. We have also experimented with an Inception-ResNet CNN backbone~\cite{szegedy2016inception}, which is an architecture integrating Inception layers~\cite{szegedy2015going} with residual connections~\cite{He2016ResNets}, which performs slightly better at the cost of increased computation.

The CNN backbone has been pre-trained for image classification on Imagenet. In all reported experiments, both the region proposal and box classifier components of the Faster-RCNN detector have been trained using only the person category in the COCO dataset and the box annotations for the remaining 79 COCO categories have been ignored. We use the Faster-RCNN implementation of \cite{huang2016speed} written in Tensorflow \cite{tensorflow2015-whitepaper}. For simplicity and to facilitate reproducibility we do not utilize multi-scale evaluation or model ensembling in the Faster-RCNN person box detection stage. Using such enhancements can further improve our results at the cost of significantly increased computation time.


\subsection{Person Pose Estimation}
\label{pose_estimation_model}

The pose estimation component of our system predicts the location of all $K=17$ person keypoints, given each person bounding box proposal delivered by the first stage.

One approach would be to use a single regressor per keypoint, as in \cite{deeppose}, but this is problematic when there is more than one person in the image patch (in which case a keypoint can occur in multiple places). A different approach addressing this issue would be to predict activation maps, as in \cite{jainiclr2014}, which allow for multiple predictions of the same keypoint. However, the size of the activation maps, and thus the localization precision, is limited by the size of the net's output feature maps, which is a fraction of the input image size, due to the use of max-pooling with decimation.

\begin{figure}[h]
\centering
\subfloat{%
  \includegraphics[width=0.32\columnwidth]{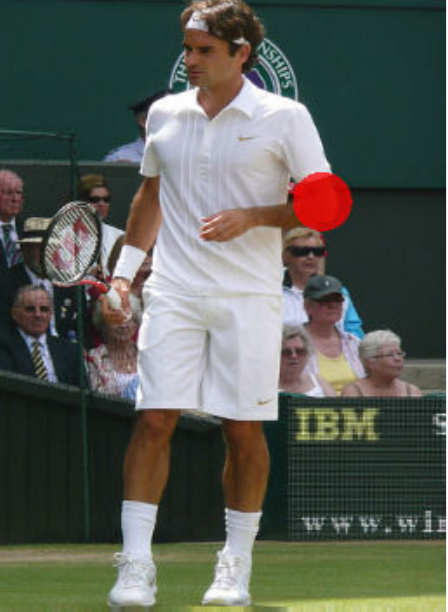}
}
\subfloat{%
  \includegraphics[width=0.32\columnwidth]{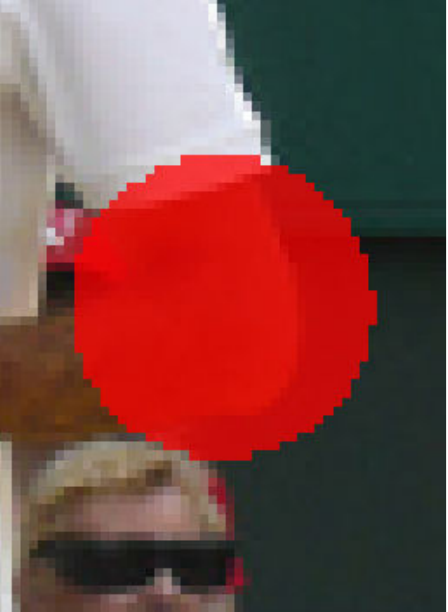}
}
\subfloat{%
  \includegraphics[width=0.32\columnwidth]{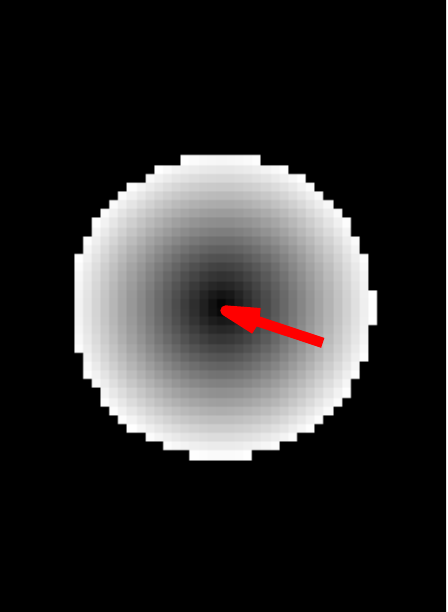}
}
\hfill
\caption{Network target outputs. \emph{Left \& Middle}: Heatmap target for the left-elbow keypoint (red indicates heatmap of 1). \emph{Right}: Offset field L2 magnitude (shown in grayscale) and 2-D offset vector shown in red).}
\label{fig:gt_heatmap_offset}
\end{figure}

In order to address the above limitations, we adopt a combined classification and regression approach. For each spatial position, we first classify whether it is in the vicinity of each of the $K$ keypoints or not (which we call a ``heatmap''), then predict a 2-D local offset vector to get a more precise estimate of the corresponding keypoint location. Note that this approach is inspired by work on object detection, where a similar setup is used to predict bounding boxes, e.g. \cite{erhan2014scalable, Ren2015}. Figure~\ref{fig:gt_heatmap_offset} illustrates these three output channels per keypoint.

\newcommand{\fcls}{f_{\mathrm{cls}}}
\newcommand{\floc}{f_{\mathrm{loc}}}

\textbf{Image Cropping}
We first make all boxes have the same fixed aspect ratio by extending either the height or the width of the boxes returned by the person detector without distorting the image aspect ratio. After that, we further enlarge the boxes to include additional image context: we use a rescaling factor equal to 1.25 during evaluation and a random rescaling factor between 1.0 and 1.5 during training (for data augmentation). We then crop from the resulting box the image and resize to a fixed crop of height 353 and width 257 pixels. We set the aspect ratio value to $353/257 = 1.37$.

\textbf{Heatmap and Offset Prediction with CNN}
We apply a ResNet with 101 layers \cite{He2016ResNets} on the cropped image in a fully convolutional fashion to produce heatmaps (one channel per keypoint) and offsets (two channels per keypoint for the x- and y- directions) for a total of $3 \cdot K$ output channels, where $K=17$ is the number of keypoints. We initialize our model from the publicly available Imagenet pretrained ResNet-101 model of \cite{He2016ResNets}, replacing its last layer with 1x1 convolution with $3 \cdot K$ outputs. We follow the approach of \cite{chen2016deeplab}: we employ atrous convolution to generate the $3 \cdot K$ predictions with an output stride of 8 pixels and bilinearly upsample them to the 353x257 crop size.

In more detail, given the image crop, let $f_k(x_i) = 1$ if the $k$-th keypoint is located at position $x_i$ and 0 otherwise. Here $k \in \{1,\ldots,K\}$ is indexing the keypoint type and $i \in \{1,\dots,N\}$ is indexing the pixel locations on the 353x257 image crop grid. Training a CNN to produce directly the highly localized activations $f_k$ (ideally delta functions) on a fine resolution spatial grid is hard.

Instead, we decompose the problem into two stages. First, for each position $x_i$ and each keypoint $k$, we compute the probability $h_k(x_i) = 1 \mbox{ if } ||x_i - l_k|| \le R$ that the point $x_i$ is within a disk of radius $R$ from the location $l_k$ of the $k$-th keypoint. We generate $K$ such heatmaps, solving a binary classification problem for each position and keypoint independently.

In addition to the heatmaps, we also predict at each position $i$ and each keypoint $k$ the 2-D offset vector $F_k(x_i) = l_k - x_i$ from the pixel to the corresponding keypoint. We generate $K$ such vector fields, solving a 2-D regression problem for each position and keypoint independently.

After generating the heatmaps and offsets, we aggregate them to produce highly localized activation maps $f_k(x_i)$ as follows:
\begin{equation}
f_k(x_i) = \sum_{j} \frac{1}{\pi R^2} G(x_j + F_k(x_j) - x_i) h_k(x_j) \,,
\end{equation}
where $G(\cdot)$ is the bilinear interpolation kernel. This is a form of Hough voting: each point $j$ in the image crop grid casts a vote with its estimate for the position of every keypoint, with the vote being weighted by the probability that it is in the disk of influence of the corresponding keypoint. The normalizing factor equals the area of the disk and ensures that if the heatmaps and offsets were perfect, then $f_k(x_i)$ would be a unit-mass delta function centered at the position of the $k$-th keypoint.

\begin{figure}
\centering
\includegraphics[width=0.5\textwidth]{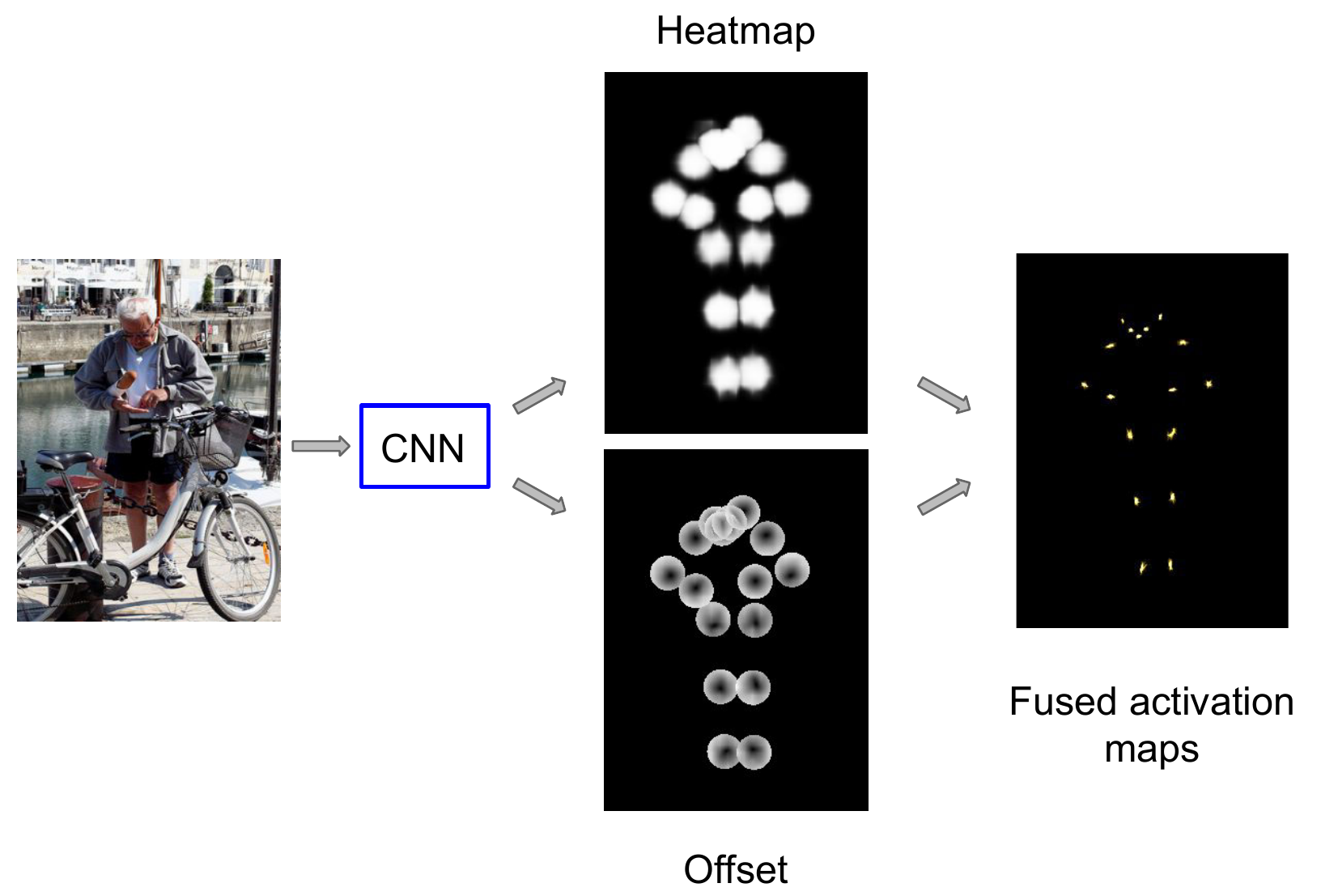}
\caption{\label{fig:heatmap_offsets}
Our fully convolutional network predicts two targets: (1) Disk-shaped heatmaps around each keypoint and (2) magnitude of the offset fields towards the exact keypoint position within the disk. Aggregating them in a weighted voting process results in highly localized activation maps. The figure shows the heatmaps and the pointwise magnitude of the offset field on a validation image. Note that in this illustration we super-impose the channels from the different keypoints.
}
\end{figure}

The process is illustrated in Figure~\ref{fig:heatmap_offsets}. We see that predicting separate heatmap and offset channels and fusing them by the proposed voting process results into highly localized activation maps which precisely pinpoint the position of the keypoints.

\textbf{Model Training}
We use a single ResNet model with two convolutional output heads. The output of the first head passes through a sigmoid function to yield the heatmap probabilities $h_k(x_i)$ for each position $x_i$ and each keypoint $k$. The training target $\bar{h}_k(x_i)$ is a map of zeros and ones, with $\bar{h}_k(x_i) = 1$ if $||x_i - l_k|| \le R$ and 0 otherwise. The corresponding loss function $L_h(\theta)$ is the sum of logistic losses for each position and keypoint separately.
To accelerate training, we follow \cite{deeper_cut} and add an extra heatmap prediction layer at  intermediate layer 50 of ResNet, which contributes a corresponding auxiliary loss term.

\eat{
To train the heatmap head, we use the following per-anchor cross entropy loss function:
\[
L^k_h(y^k, \calI, \theta) = - \sum_{a \in \calA}
  t_a^k \log p(h_a^k = 1 | \calI, \theta)
  \]
where $y^k \in \calX$ is the true location of keypoint $k$,
and $t_a^k  = \delta_r(y^k, a)$ encodes the true location
using a heatmap with a disk of radius $r$
centered on the true location.
If the keypoint is not visible, then $t_a^k=0$ for all locations $a$,
so the model will be encouraged not to hallucinate invisible joints.
(In some settings, we might want the model to perform amodal completion,
but that is beyond the scope of this paper.)
}

For training the offset regression head, we penalize the difference between the predicted and ground truth offsets. The corresponding loss is
\begin{equation}
L_o(\theta) = \sum_{k=1:K} \sum_{i : ||l_k - x_i|| \le R} H(||F_k(x_i) - (l_k - x_i)||) \,,
\end{equation}
where $H(u)$ is the Huber robust loss, $l_k$ is the position of the $k$-th keypoint, and we only compute the loss for positions $x_i$ within a disk of radius $R$ from each keypoint \cite{Ren2015}.

The final loss function has the form
\begin{equation}
\label{pose_loss}
L(\theta) = \lambda_h L_h(\theta) + \lambda_o L_o(\theta) \,,
\end{equation}
where $\lambda_h=4$ and $\lambda_o=1$ is a scalar factor to balance the loss function terms. We sum this loss over all the images in a minibatch, and then apply stochastic gradient descent.

An important consideration in model training is how to treat cases where multiple people exist in the image crop in the computation of heatmap loss. When computing the heatmap loss at the intermediate layer, we exclude contributions from within the disks around the keypoints of background people. When computing the heatmap loss at the final layer, we treat as positives only the disks around the keypoints of the foreground person and as negatives everything else, forcing the model to predict correctly the keypoints of the person in the center of the box.

\paragraph{Pose Rescoring}
At test time,  we apply the model to each image crop. Rather than just relying on the confidence from the person detector, we compute a refined confidence estimate, which takes into account the confidence of each keypoint. In particular, we maximize over locations and average over keypoints, yielding our final instance-level pose detection score:
\begin{equation}
\label{pose_score}
\mbox{score}(\calI) = \frac{1}{K} \sum_{k=1}^K \max_{x_i} f_k(x_i)
\end{equation}
We have found that ranking our system's pose estimation proposals using~\ref{pose_score} significantly improves AP compared to using the score delivered by the Faster-RCNN box detector.

\paragraph{OKS-Based Non Maximum Suppression}
Following standard practice, we use non maximal suppression (NMS) to eliminate multiple detections in the person-detector stage. The standard approach measures overlap using intersection over union (IoU) of the boxes. We propose a more refined variant which takes the keypoints into account. In particular, we measure overlap using the object keypoint similarity (OKS) for two candidate pose detections. Typically, we use a relatively high IOU-NMS threshold (0.6 in our experiments) at the output of the person box detector to filter highly overlapping boxes. The subtler OKS-NMS at the output of the pose estimator is better suited to determine if two candidate detections correspond to false positives (double detection of the same person) or are true positives (two people in close proximity to each other).

\eat{
 
\subsection{Complete system}

Since our pose model is fully convolutional, it is possible to apply it only once to the whole image,
and then to derive the predicted keypoint locations
and scores for each person proposal by cropping the relevant heatmap and offset map,
rather than running the model on each crop separately.
Furthermore, by using a differentiable cropping procedure,
as in \cite{girshick2015fast}, we can perform end-to-end fine tuning of the whole system, all the way back to the features of the person detector. However, we have not yet implemented this. Instead, we first run the person detector (which is also fully convolutional), and then apply the pose estimator to each predicted person box.

Following standard practice, we use non maximal suppression (NMS) to eliminate detections that overlap too closely. The standard approach measures overlap using intersection over union (IoU) of the boxes.
We propose a more refined variant, that takes the keypoints into account.
In particular, we measure the OKS score \cite{} between corresponding keypoints for two candidate detections, and average this over all keypoints. We then use this as a measure of overlap.
\todo{Is this correct?}

}

\section{Experimental Evaluation}
\label{sec:results}

\newlength{\myheightA}
\setlength{\myheightA}{4.08cm}
\newlength{\myheightB}
\setlength{\myheightB}{3.7cm}
\newlength{\myheightC}
\setlength{\myheightC}{3.625cm}
\newlength{\myheightD}
\setlength{\myheightD}{3.8cm}
\newlength{\myheightE}
\setlength{\myheightE}{3.82cm}
\newlength{\myheightF}
\setlength{\myheightF}{3cm}

\begin{figure*}
  \begin{center}
      \centering
  \adjustbox{height=\myheightA}
      {\includegraphics[trim={1.6cm 0 1.0cm 0},clip]{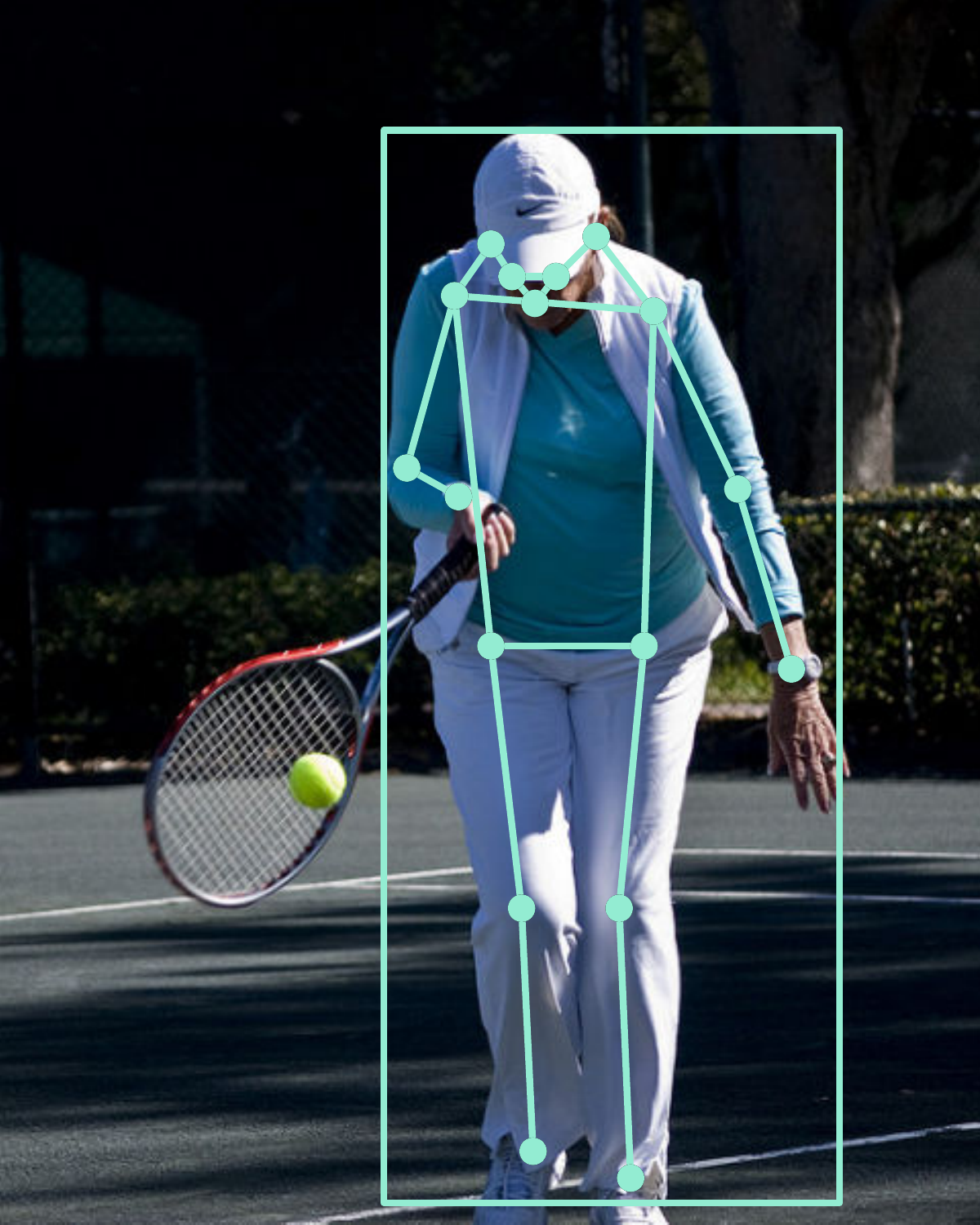}}
  \adjustbox{height=\myheightA}
      {\includegraphics[trim={6cm 0 2.5cm 0},clip]{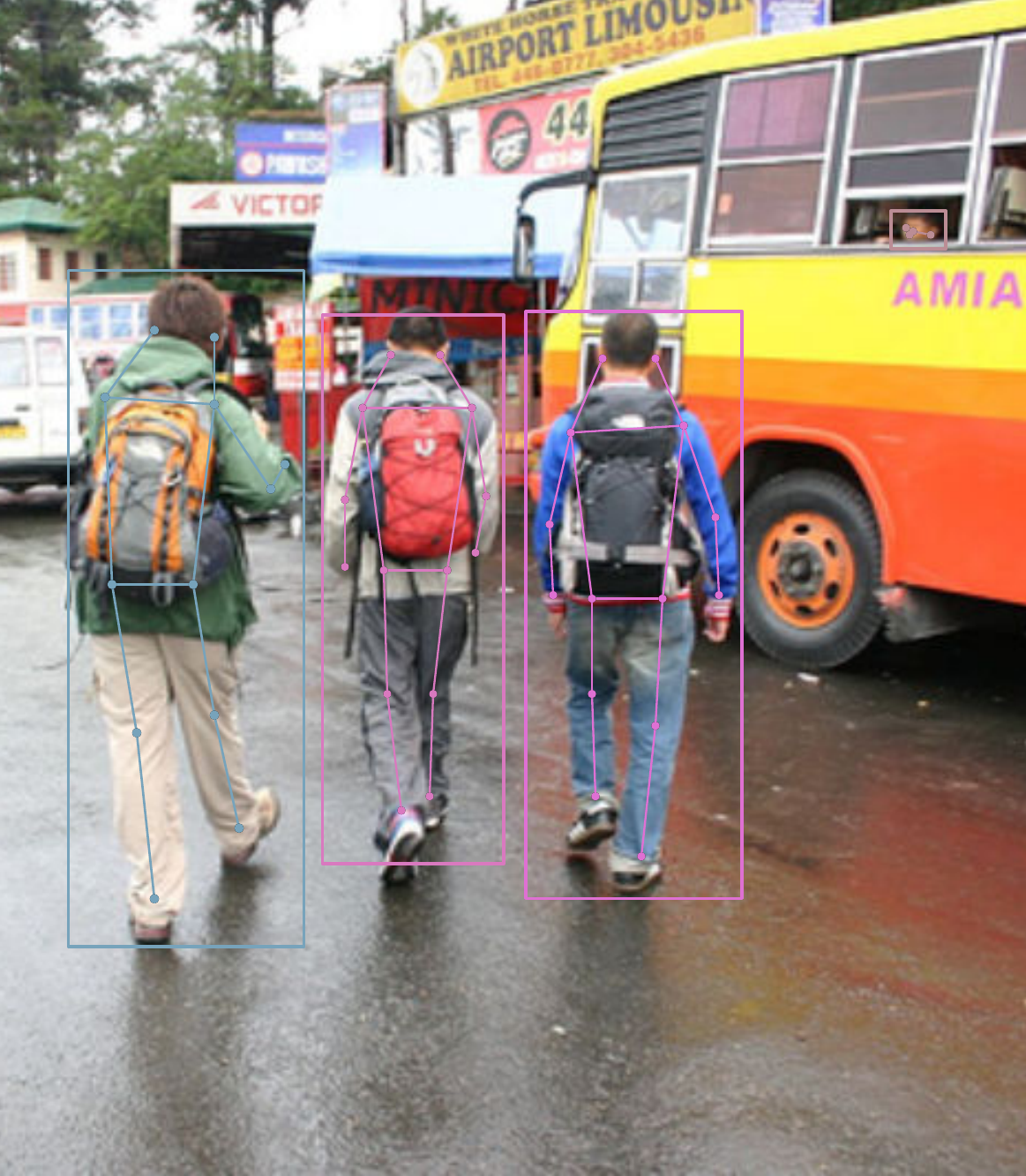}}
  \adjustbox{height=\myheightA}
      {\includegraphics[trim={10cm 0 3cm 0},clip]{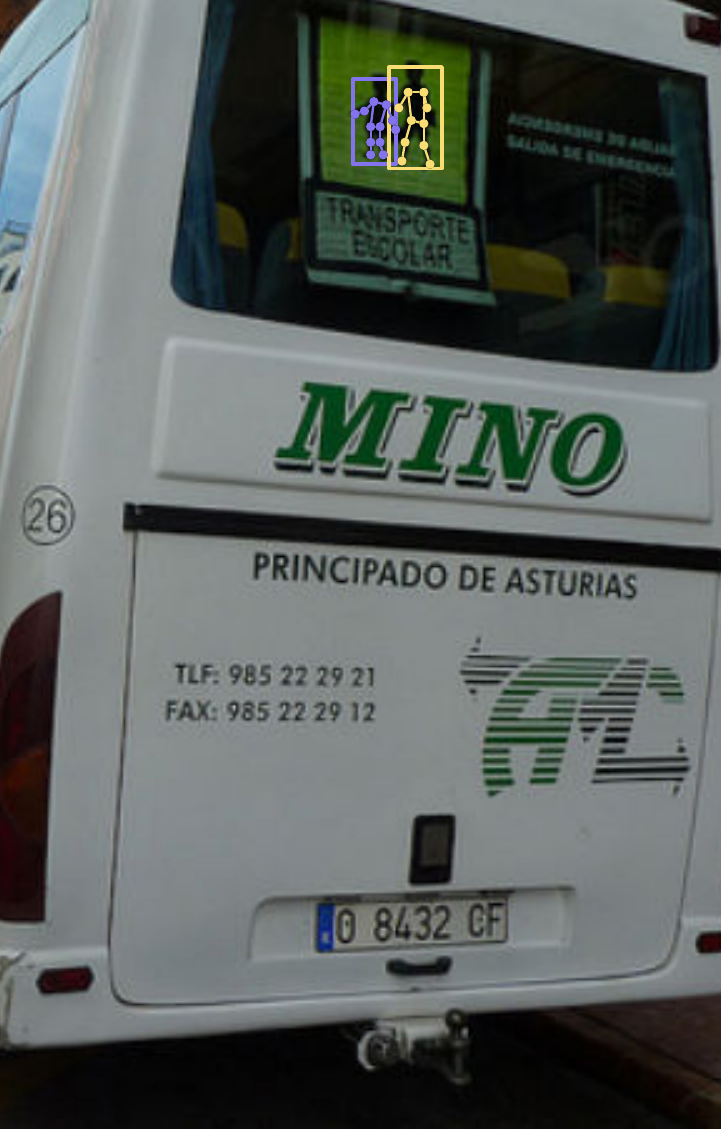}} 
  \adjustbox{height=\myheightA}
      {\includegraphics[trim={4cm 0 4.5cm 0cm},clip]{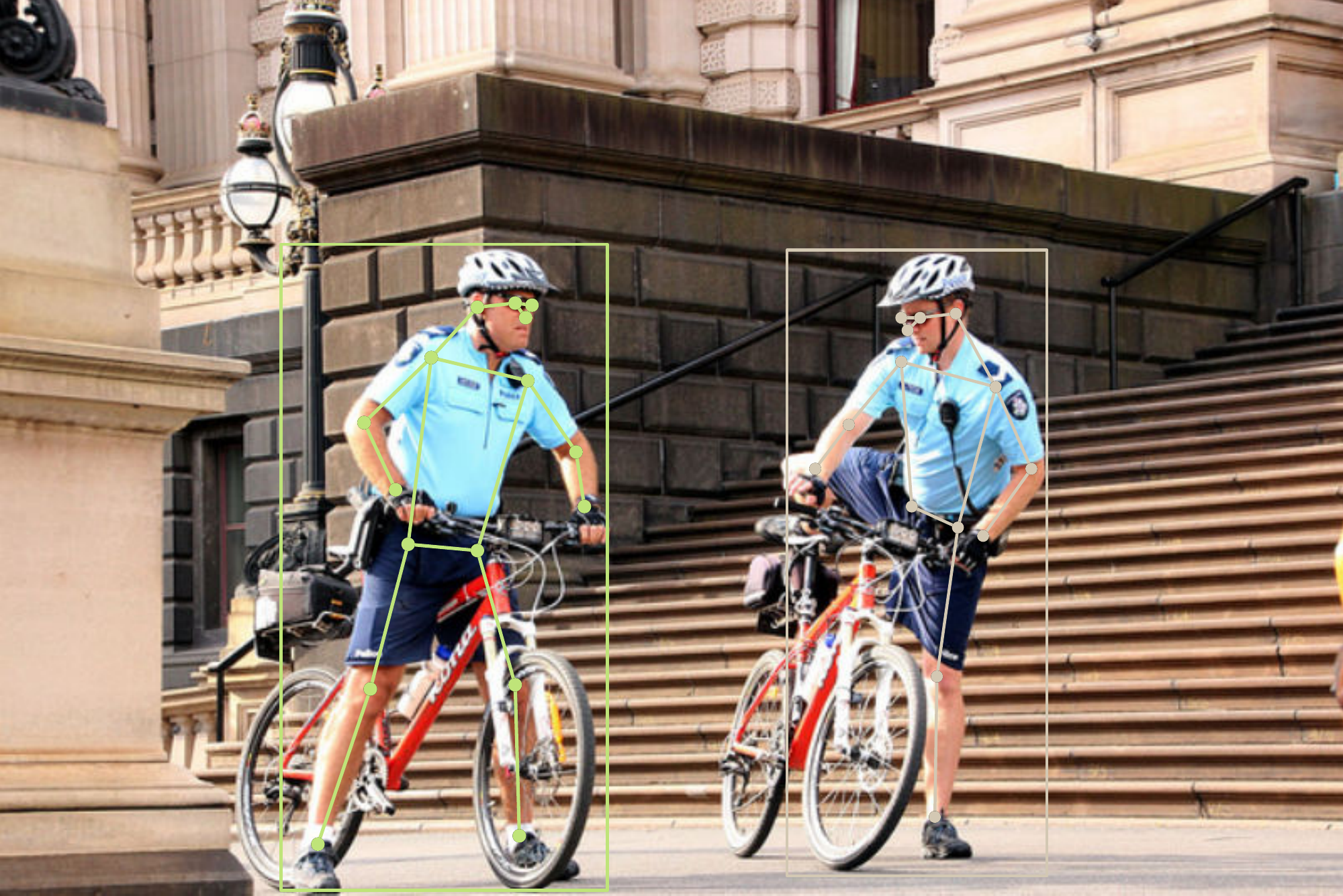}} 
  \adjustbox{height=\myheightA}
      {\includegraphics[trim={0cm 0 0cm 0},clip]{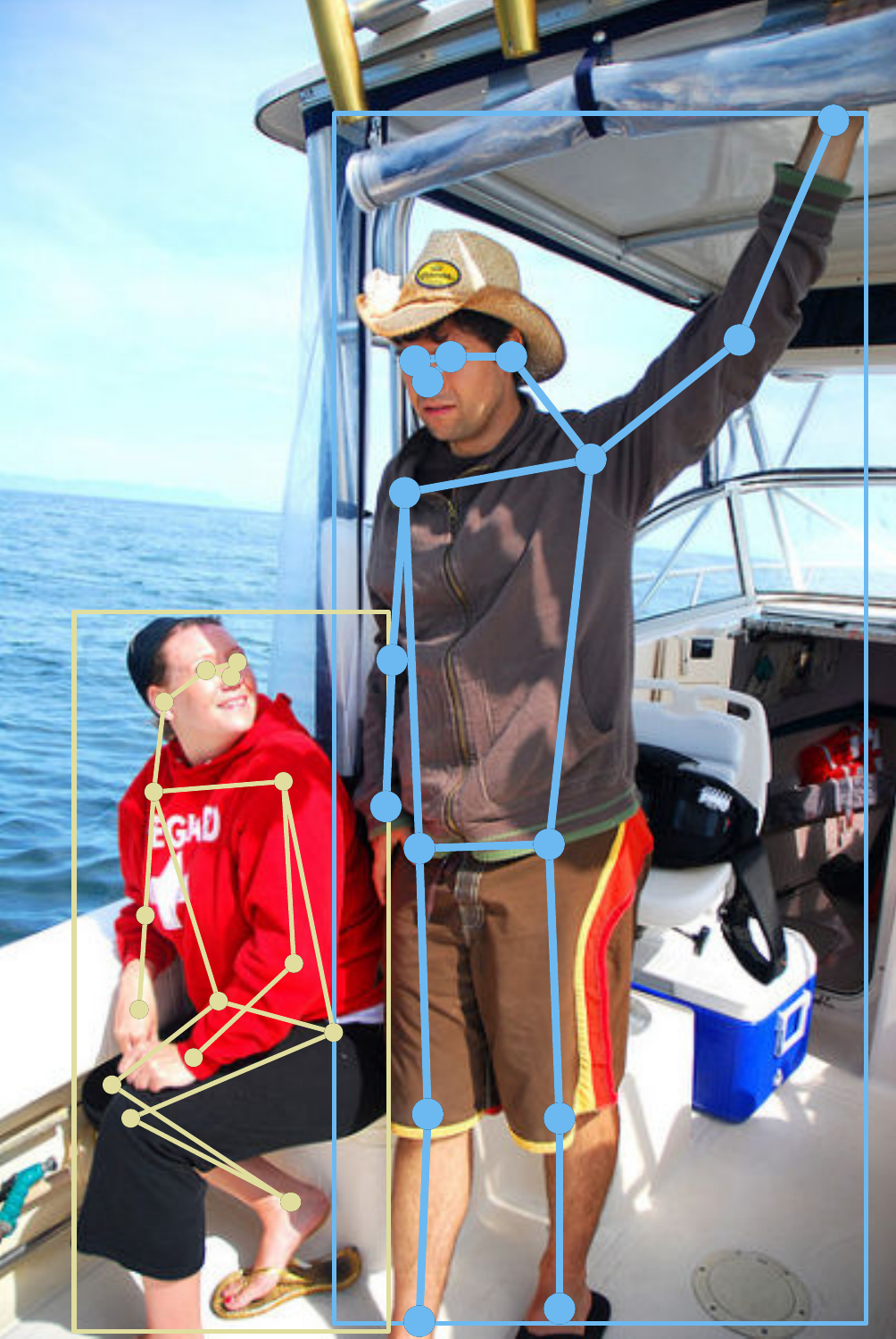}}
   \adjustbox{height=\myheightA}
      {\includegraphics[trim={4cm 0 0cm 0},clip]{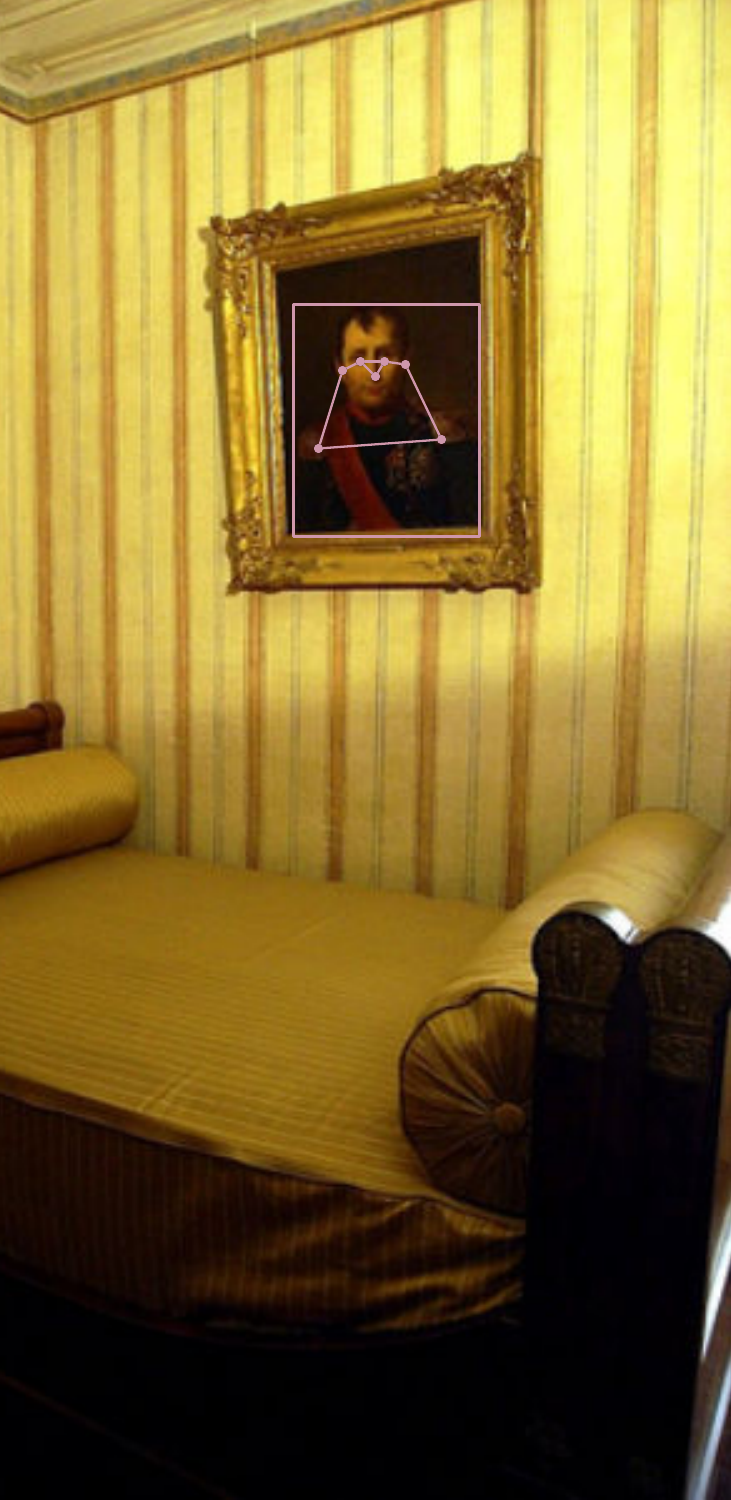}} 

  \vspace{1mm}
  \centering
  \adjustbox{height=\myheightB}
      {\includegraphics[trim={8cm 0 0cm 0},clip]{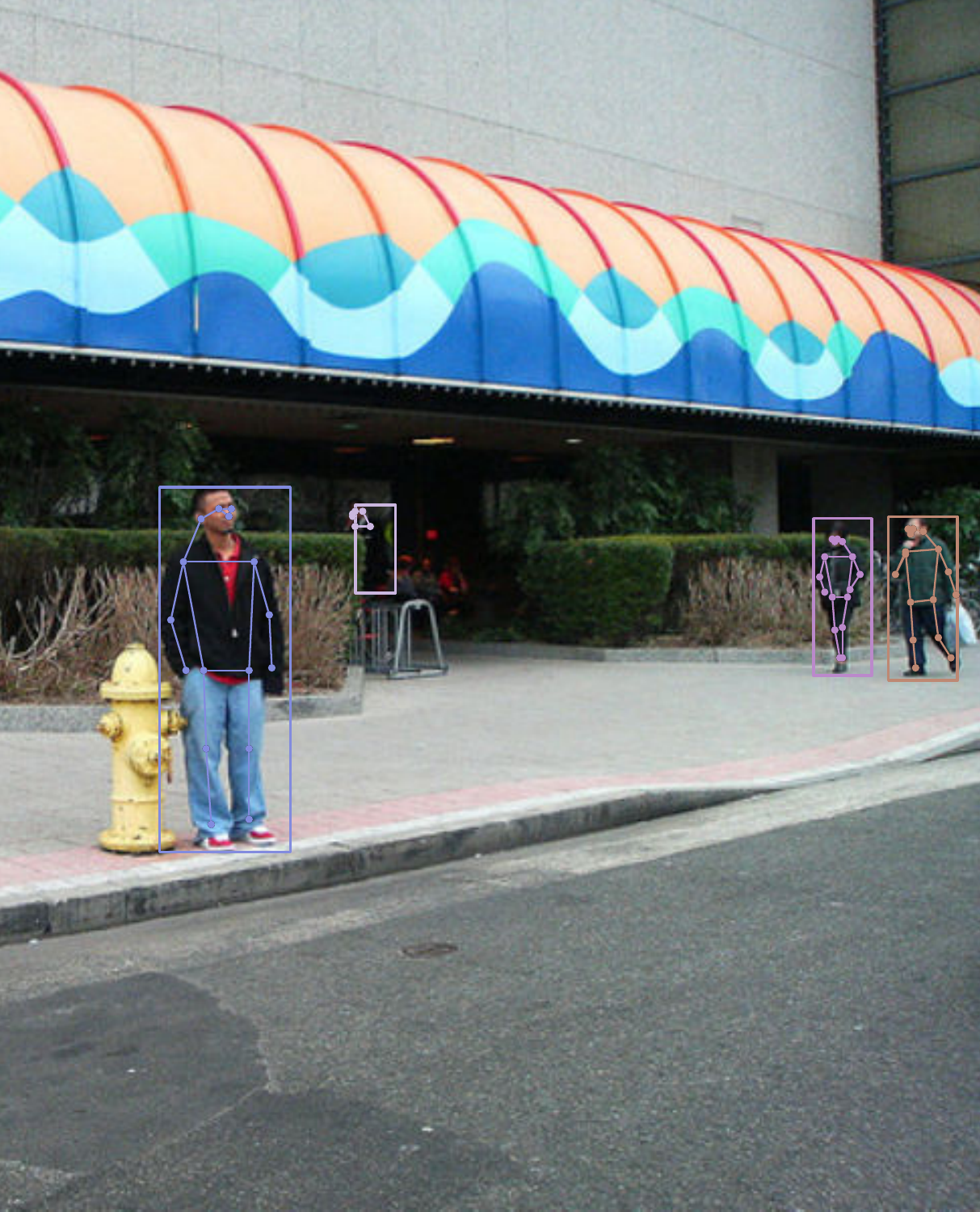}}      
  \adjustbox{height=\myheightB}
      {\includegraphics[trim={0cm 0 0cm 4cm},clip]{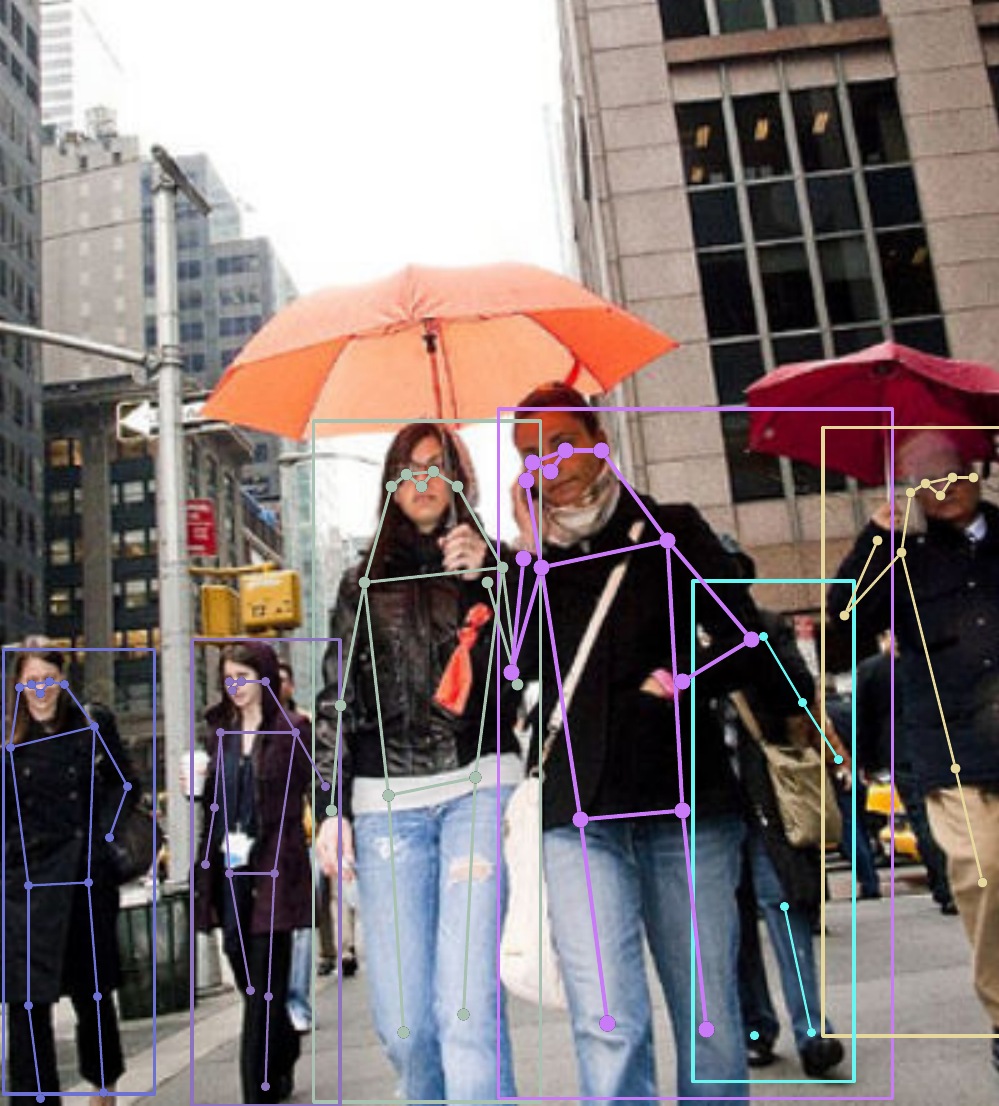}}
  \adjustbox{height=\myheightB}
      {\includegraphics[trim={1cm 0 2cm 0},clip]{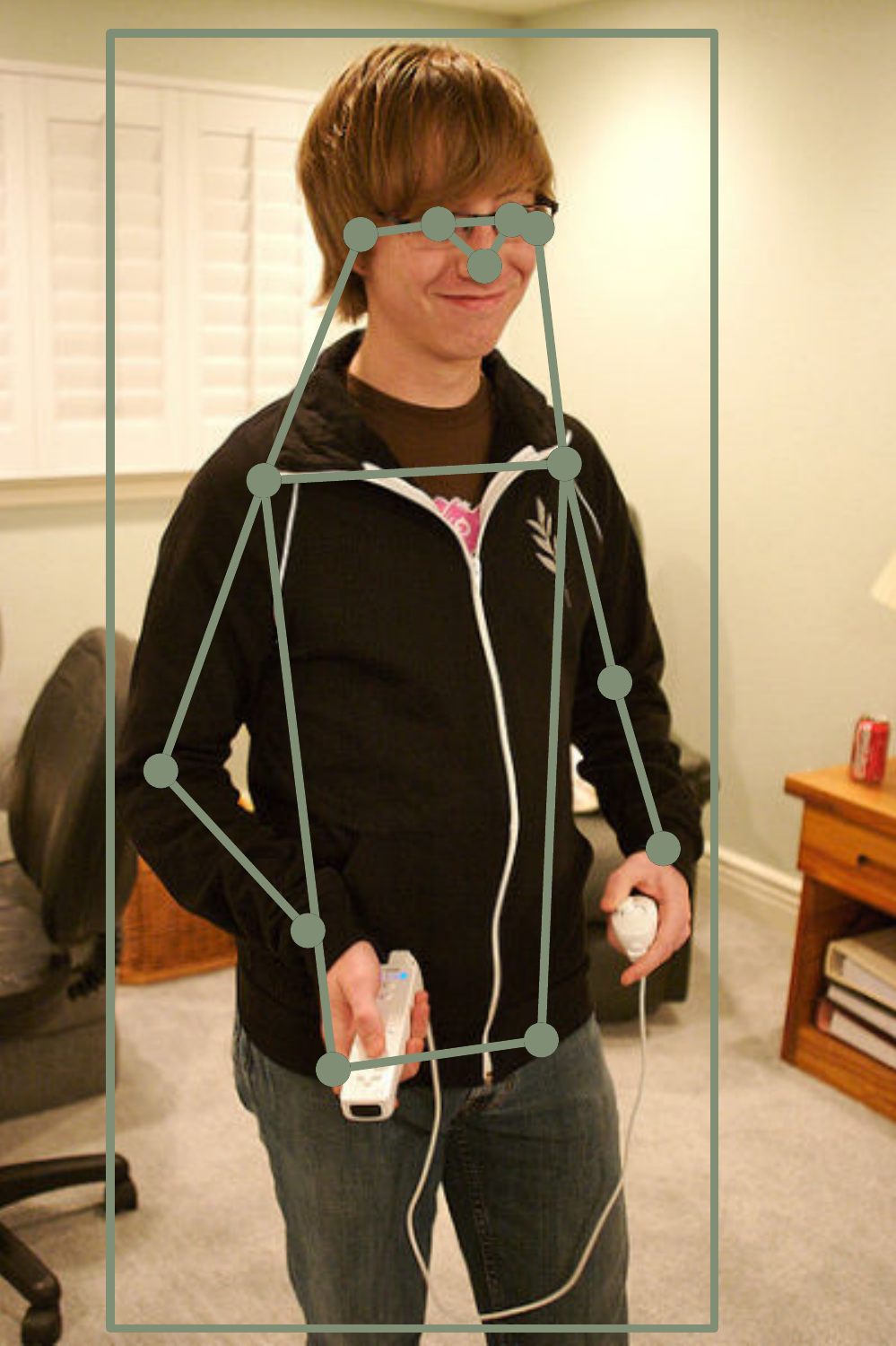}}
  \adjustbox{height=\myheightB}
      {\includegraphics[trim={8cm 0 0cm 0},clip]{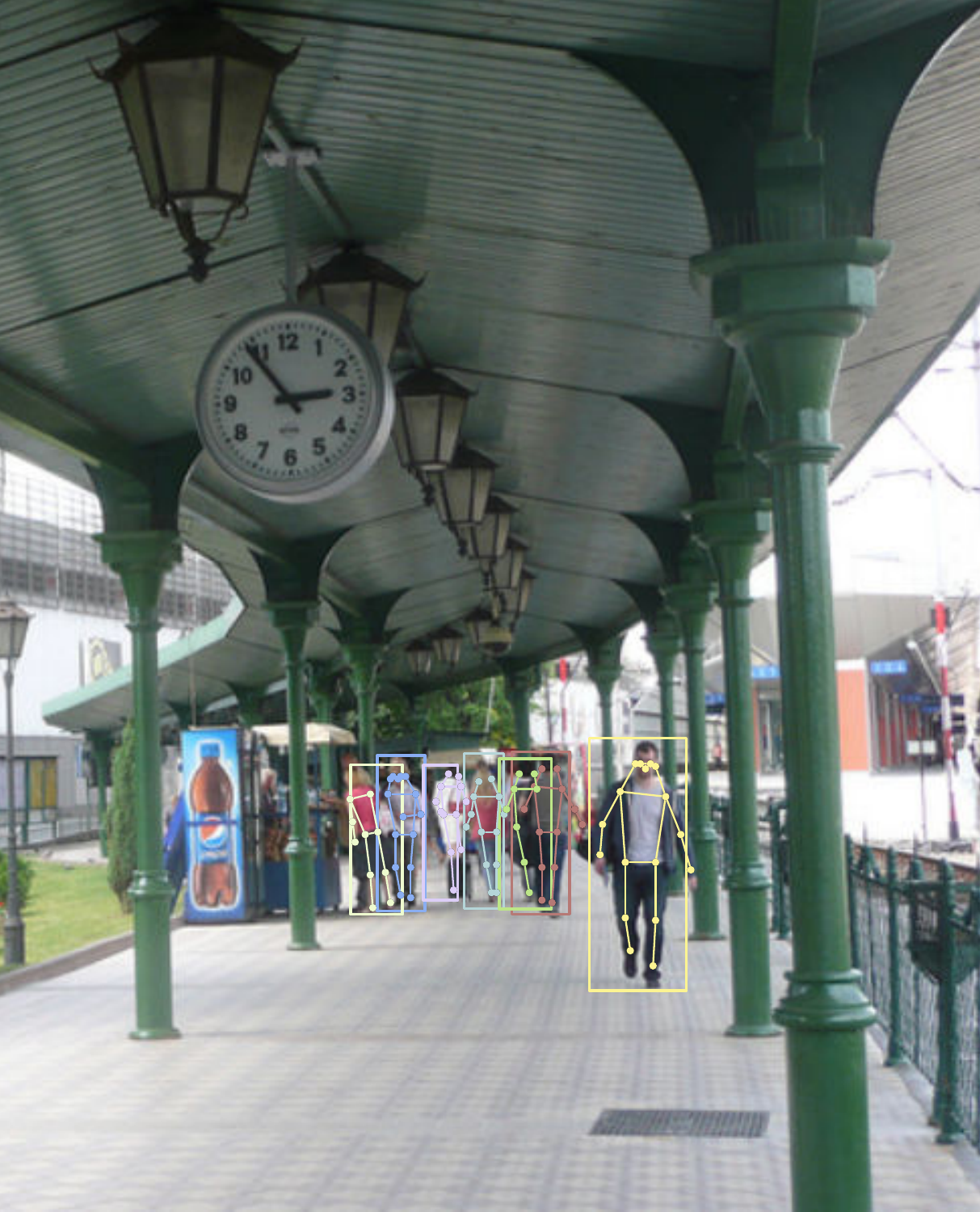}}
  \adjustbox{height=\myheightB}
      {\includegraphics[trim={1cm 0 1.5cm 0},clip]{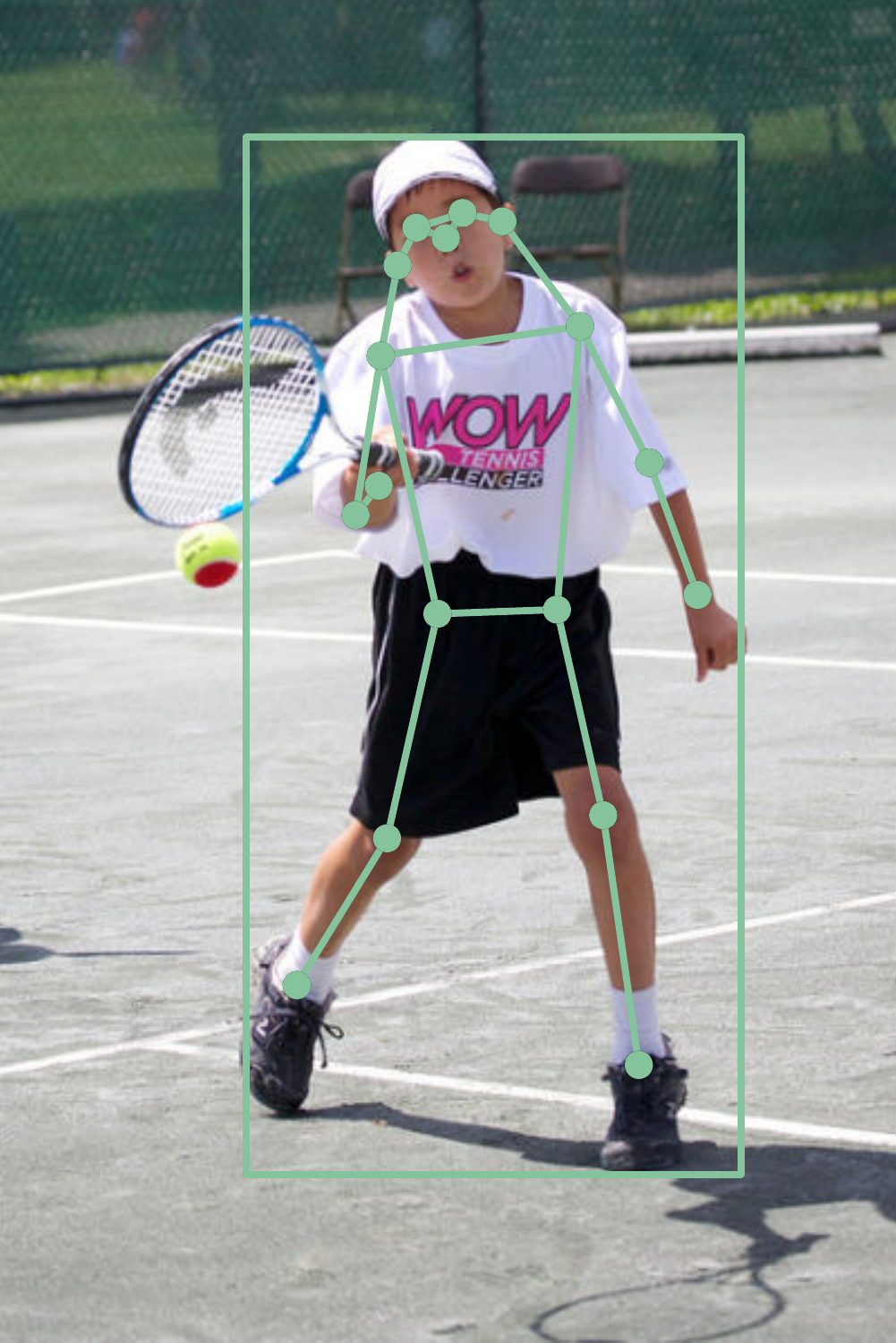}}
  \adjustbox{height=\myheightB}
      {\includegraphics[trim={7cm 0 4cm 0},clip]{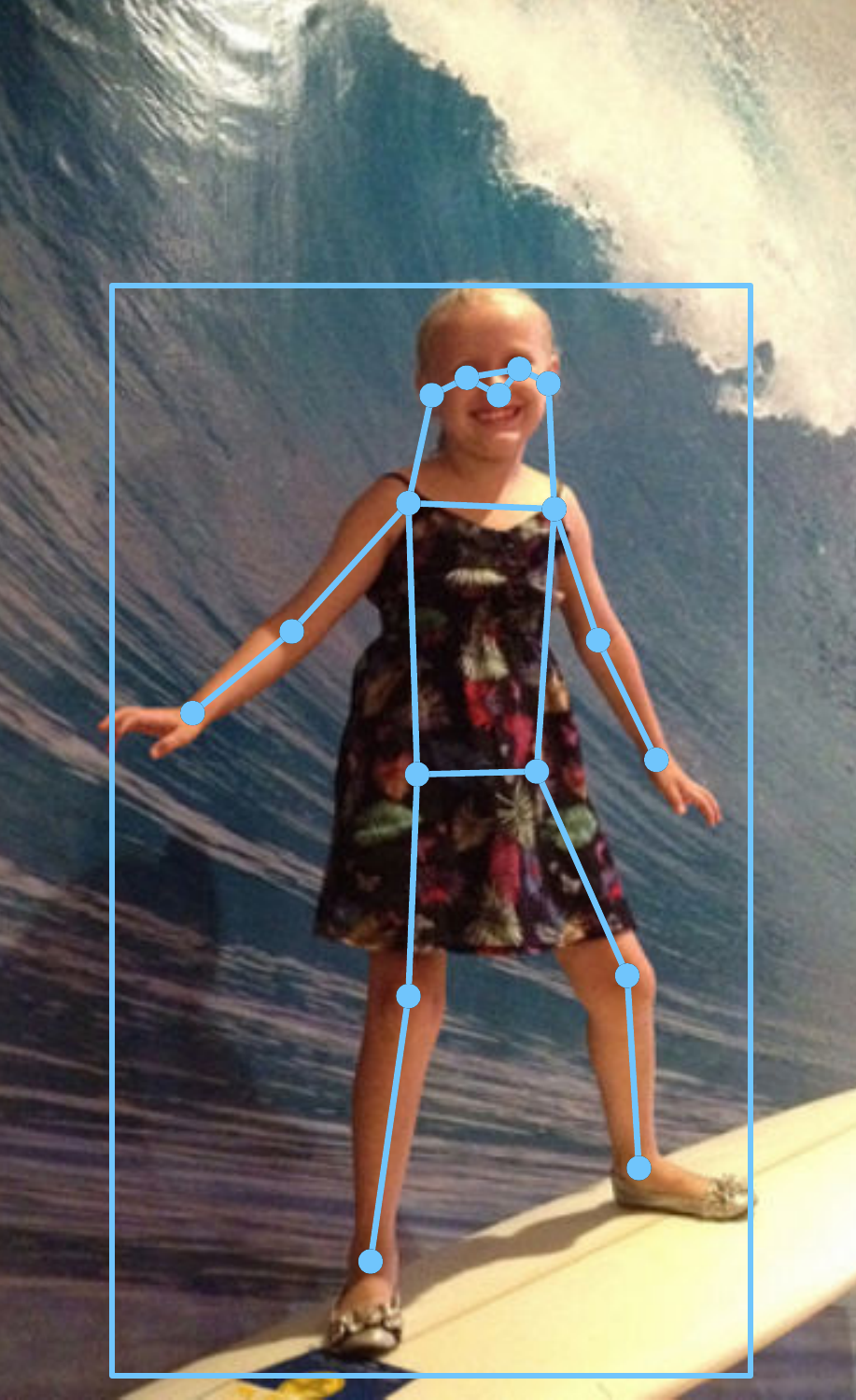}}
  \adjustbox{height=\myheightB}
      {\includegraphics[trim={3cm 0 10cm 0},clip]{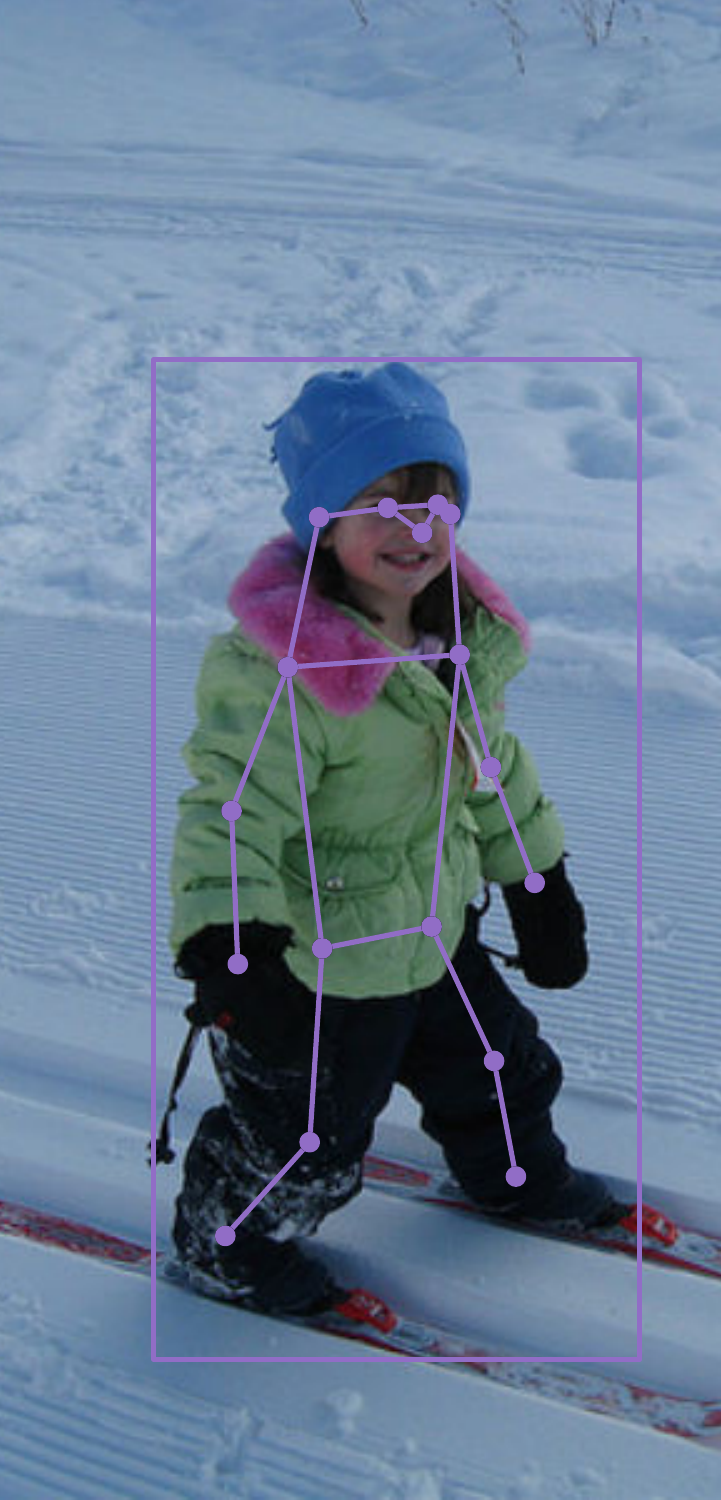}}      
      
  \vspace{1mm}
  \centering
  \adjustbox{height=\myheightC}
      {\includegraphics[trim={7cm 0 5cm 0},clip]{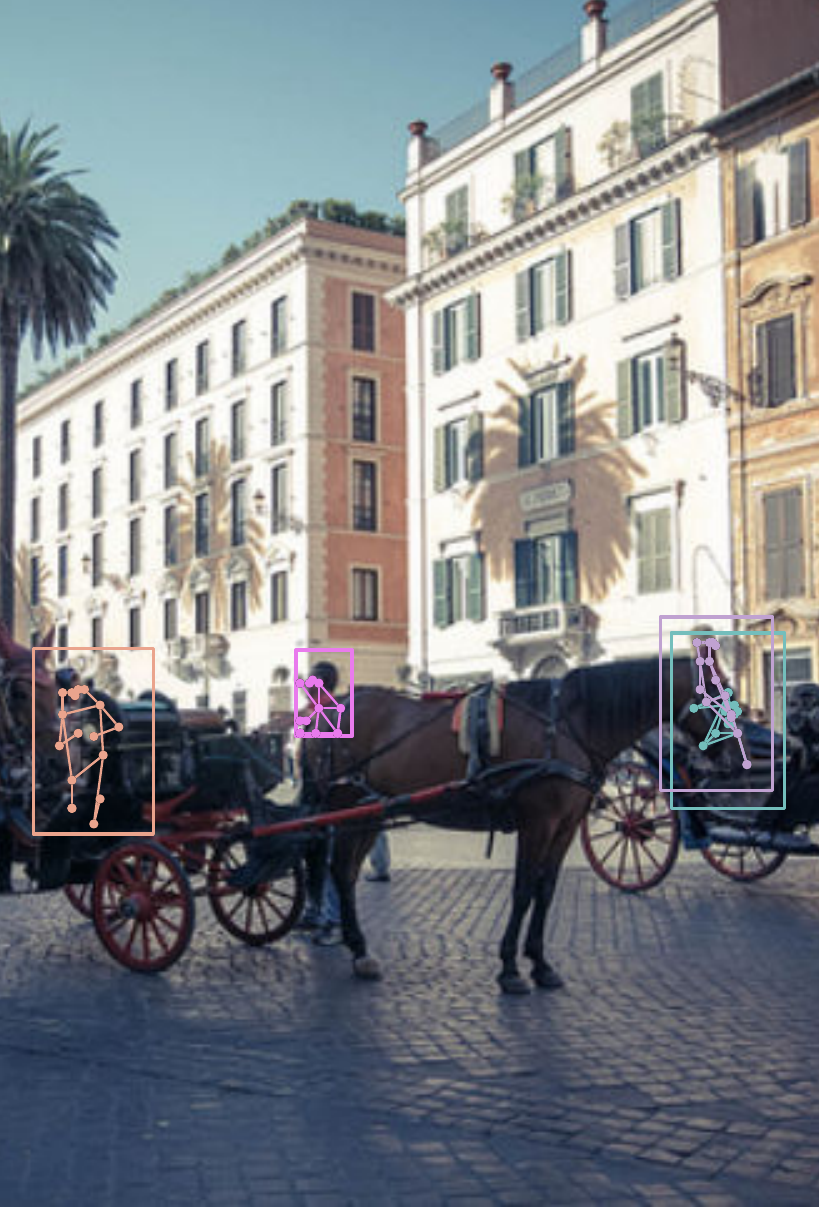}}
  \adjustbox{height=\myheightC}
      {\includegraphics[trim={0cm 0 5cm 0},clip]{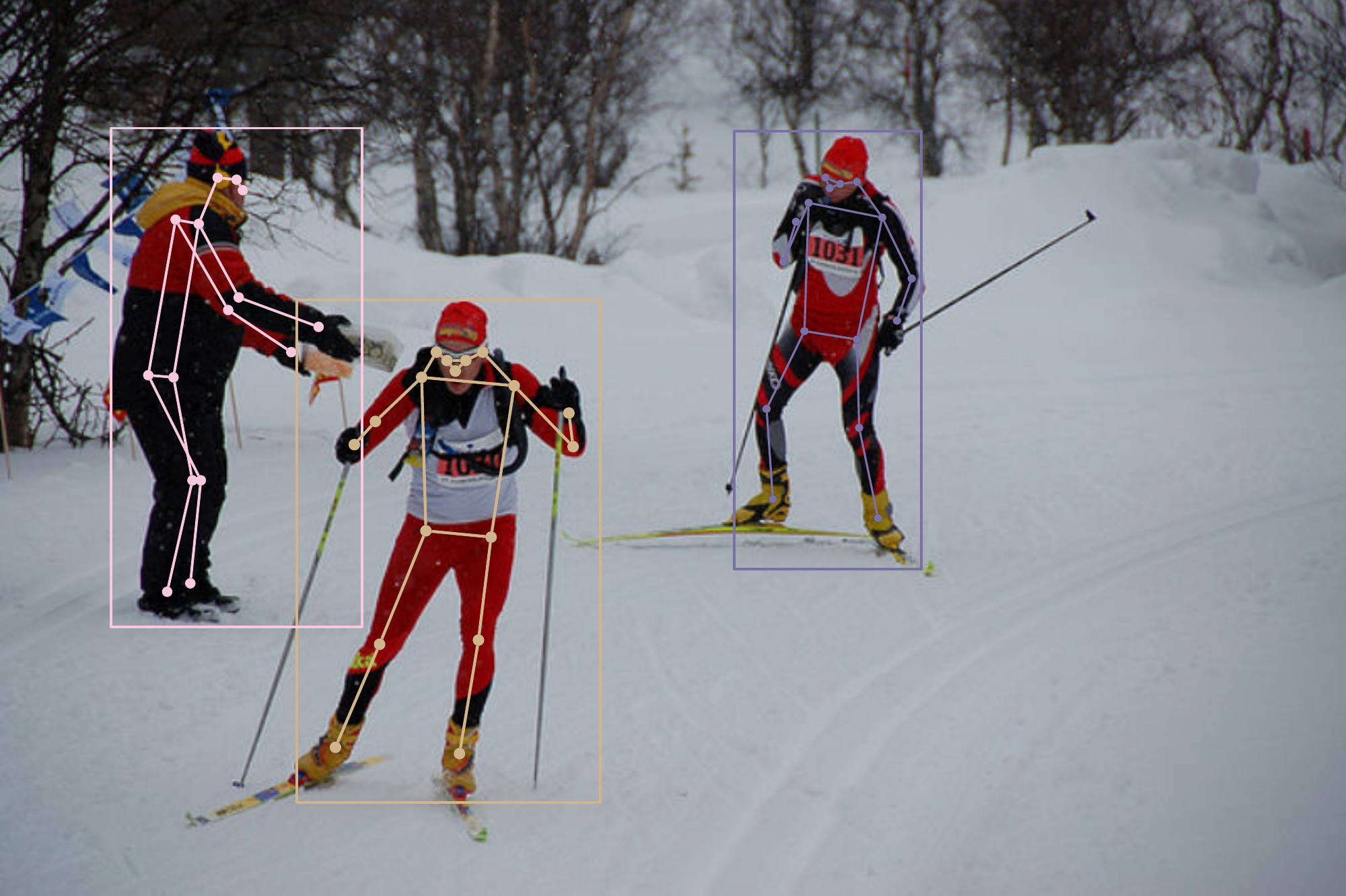}}      
   \adjustbox{height=\myheightC}
      {\includegraphics[trim={0cm 0 2cm 0},clip]{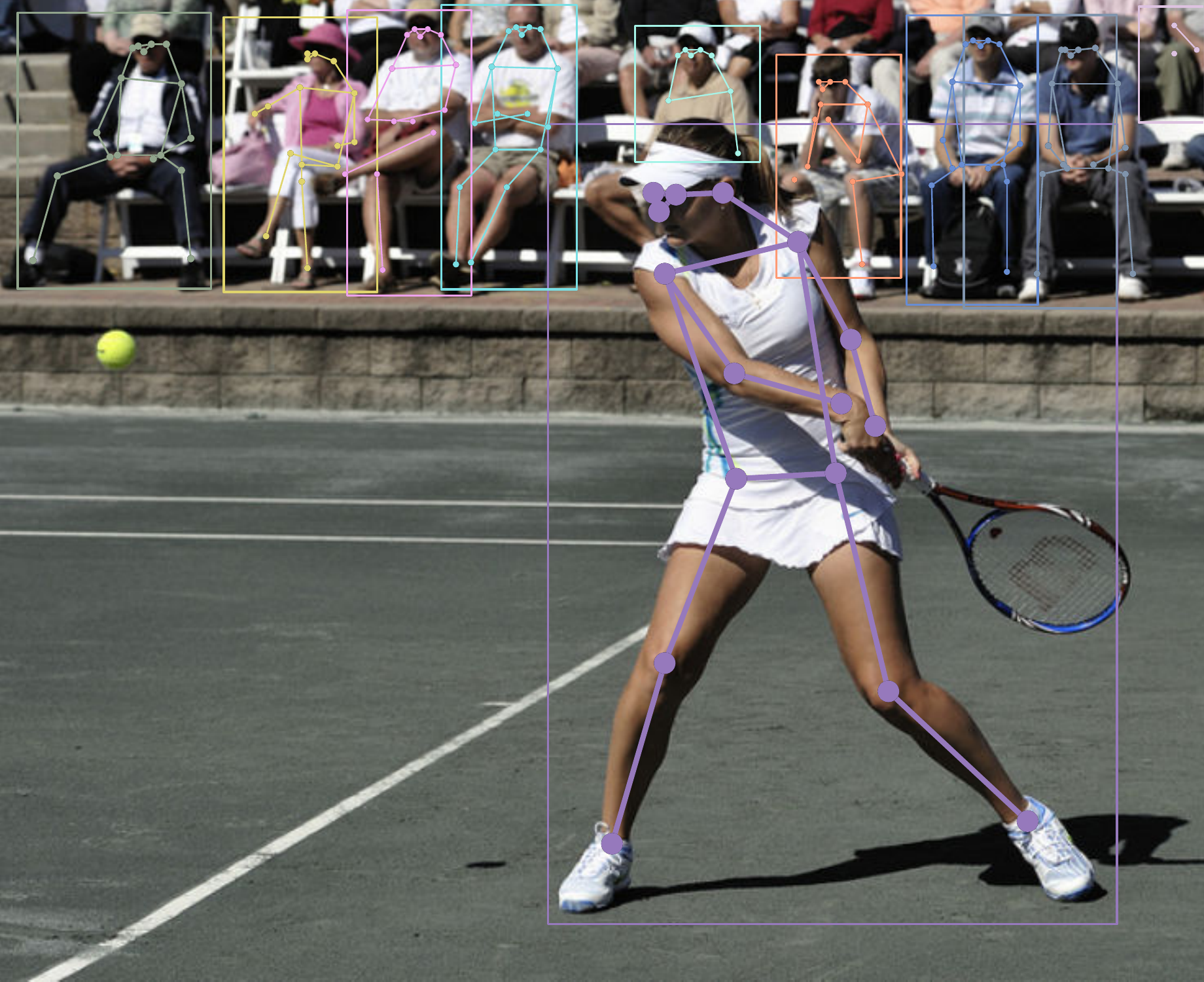}}        
   \adjustbox{height=\myheightC}
      {\includegraphics[trim={0cm 0 13cm 0},clip]{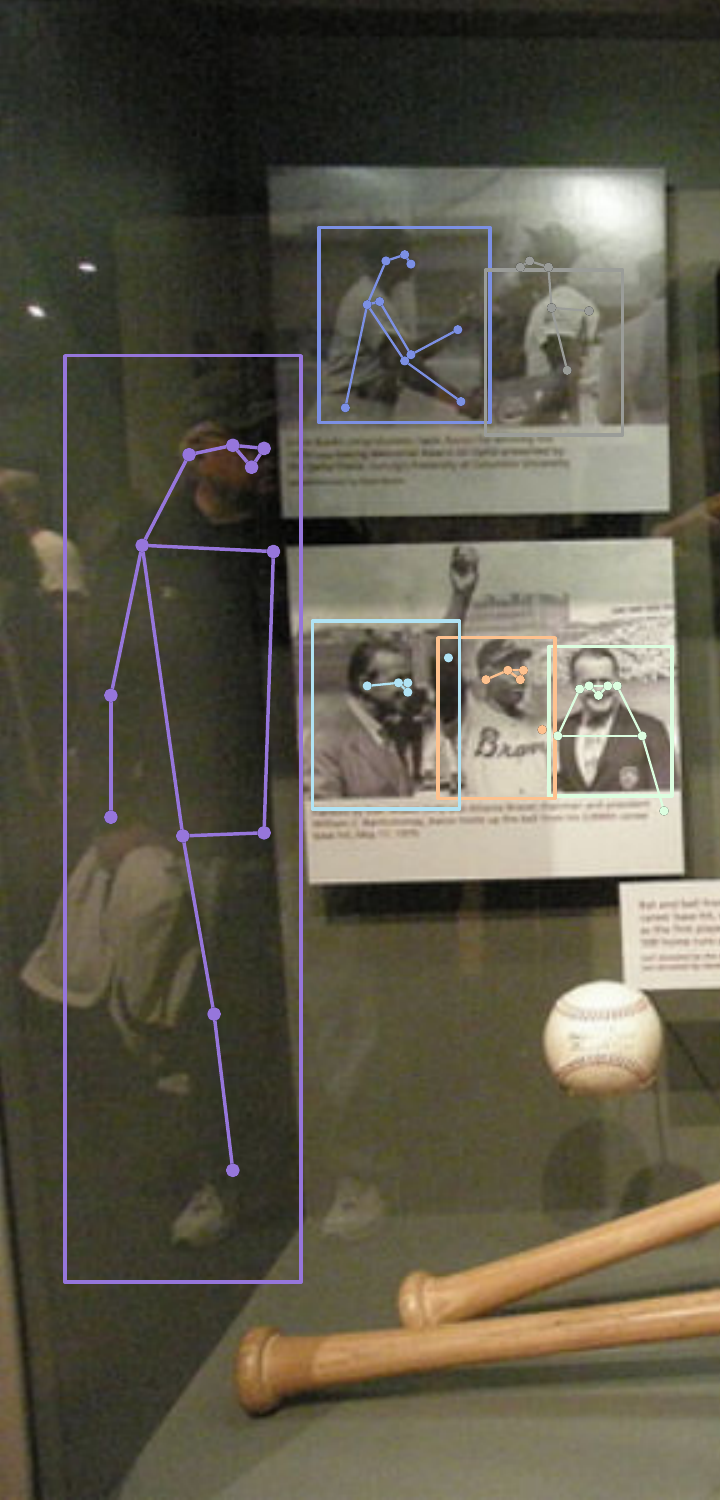}}    
   \adjustbox{height=\myheightC}
      {\includegraphics[trim={0cm 0 0cm 0},clip]{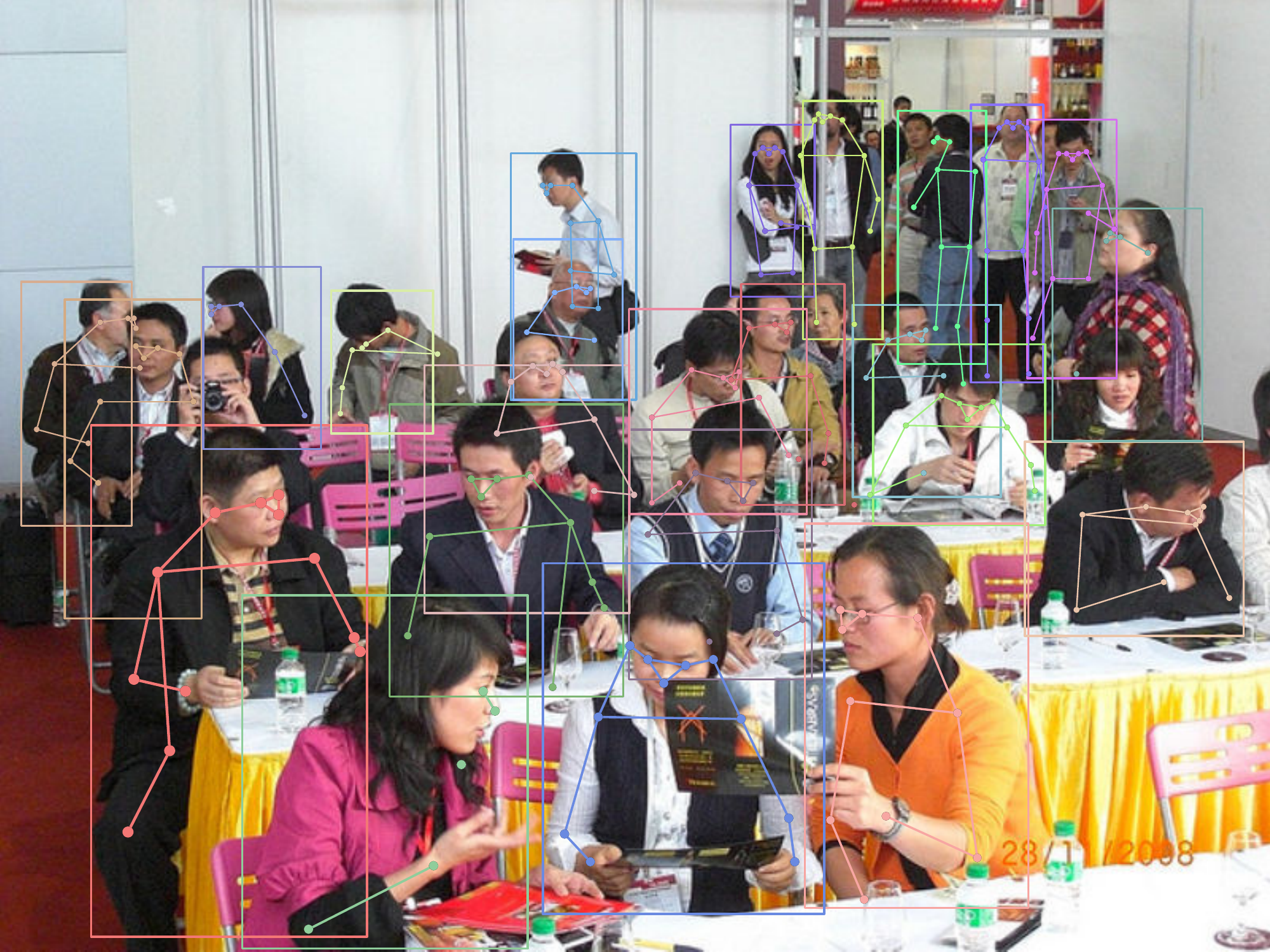}}    
      
  \vspace{1mm}
  \centering
  \adjustbox{height=\myheightD}
      {\includegraphics[trim={0cm 0 0cm 0},clip]{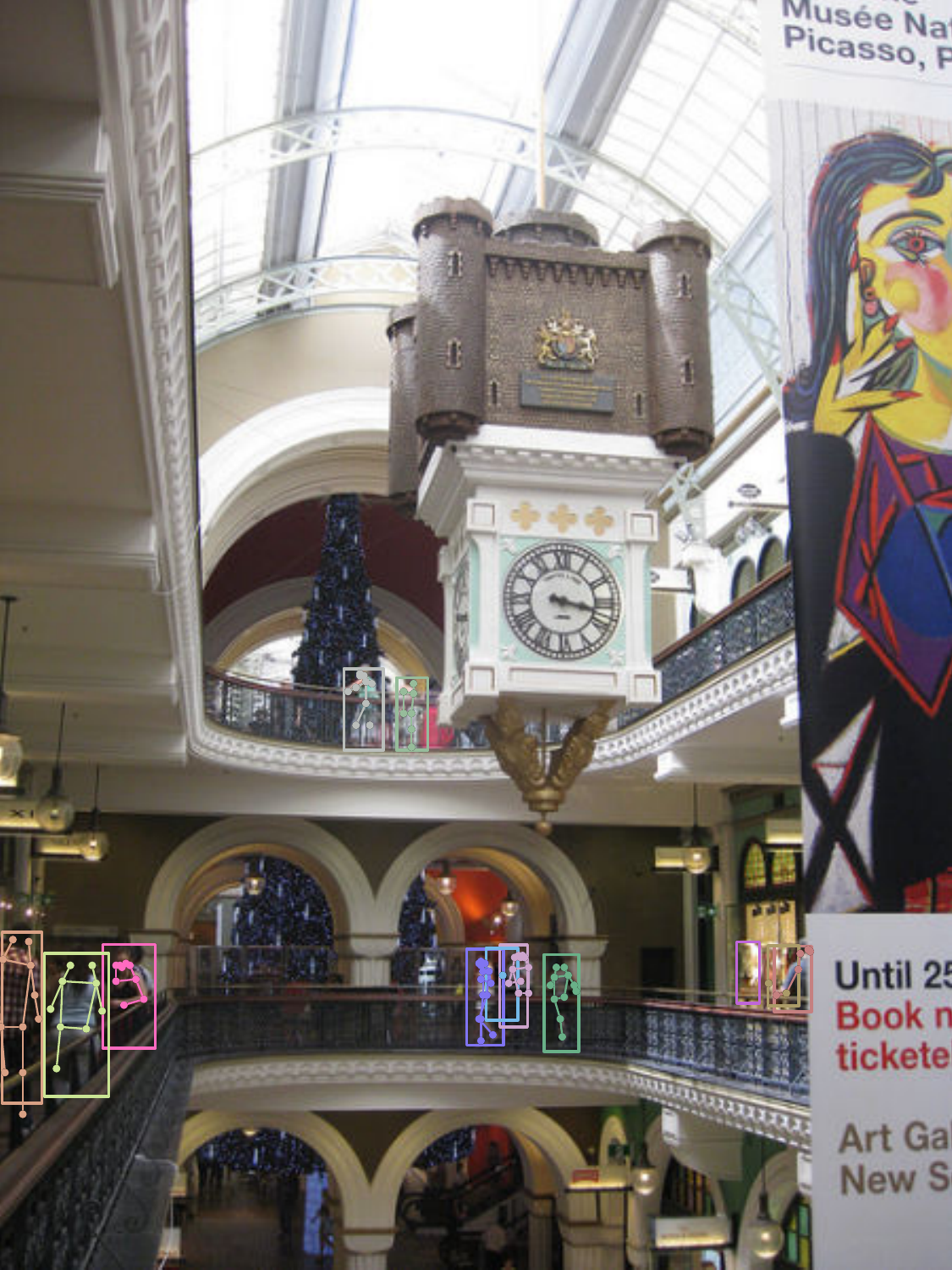}} 
  \adjustbox{height=\myheightD}
      {\includegraphics[trim={7cm 0 1cm 0},clip]{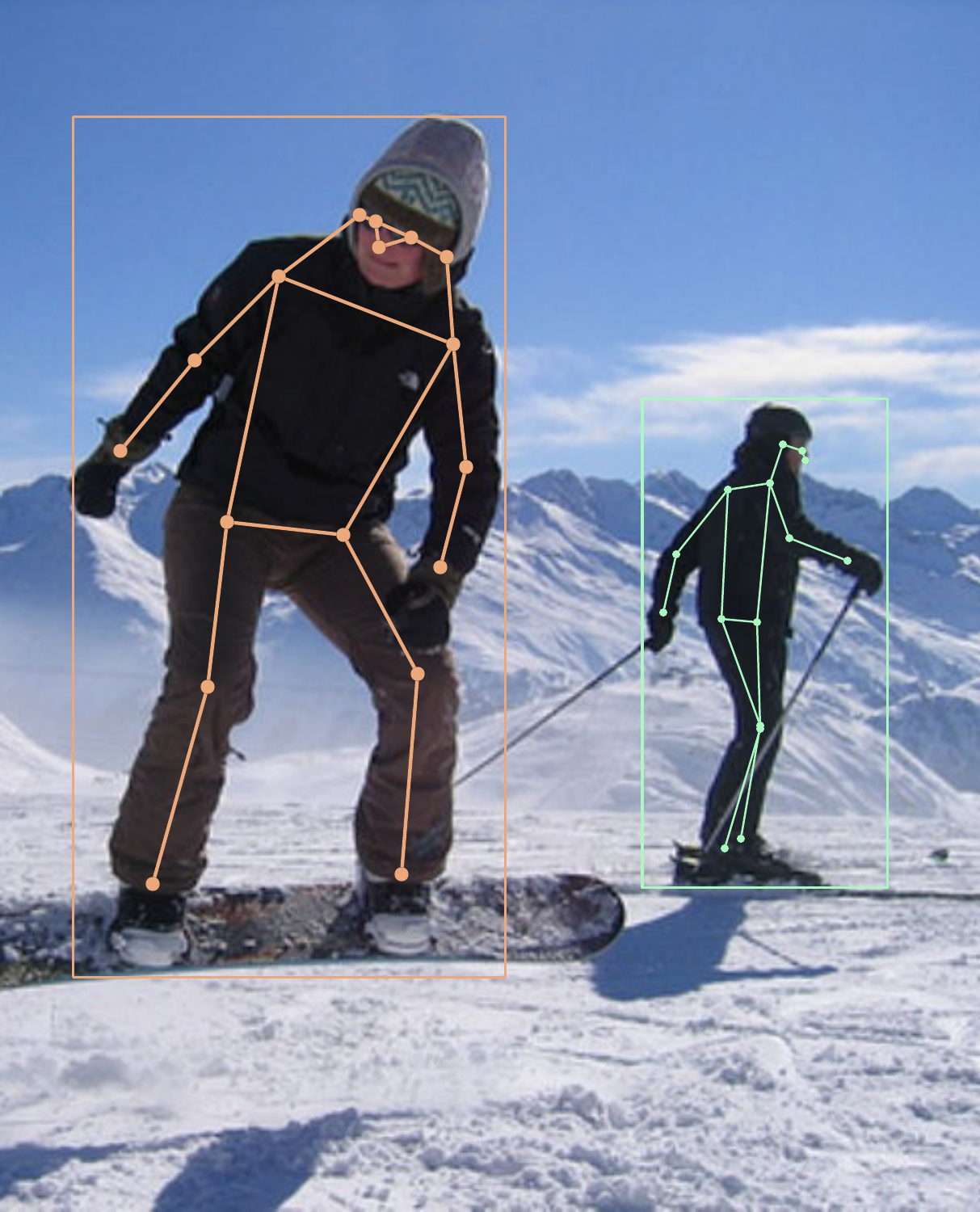}} 
  \adjustbox{height=\myheightD}
      {\includegraphics[trim={0cm 0 14cm 0},clip]{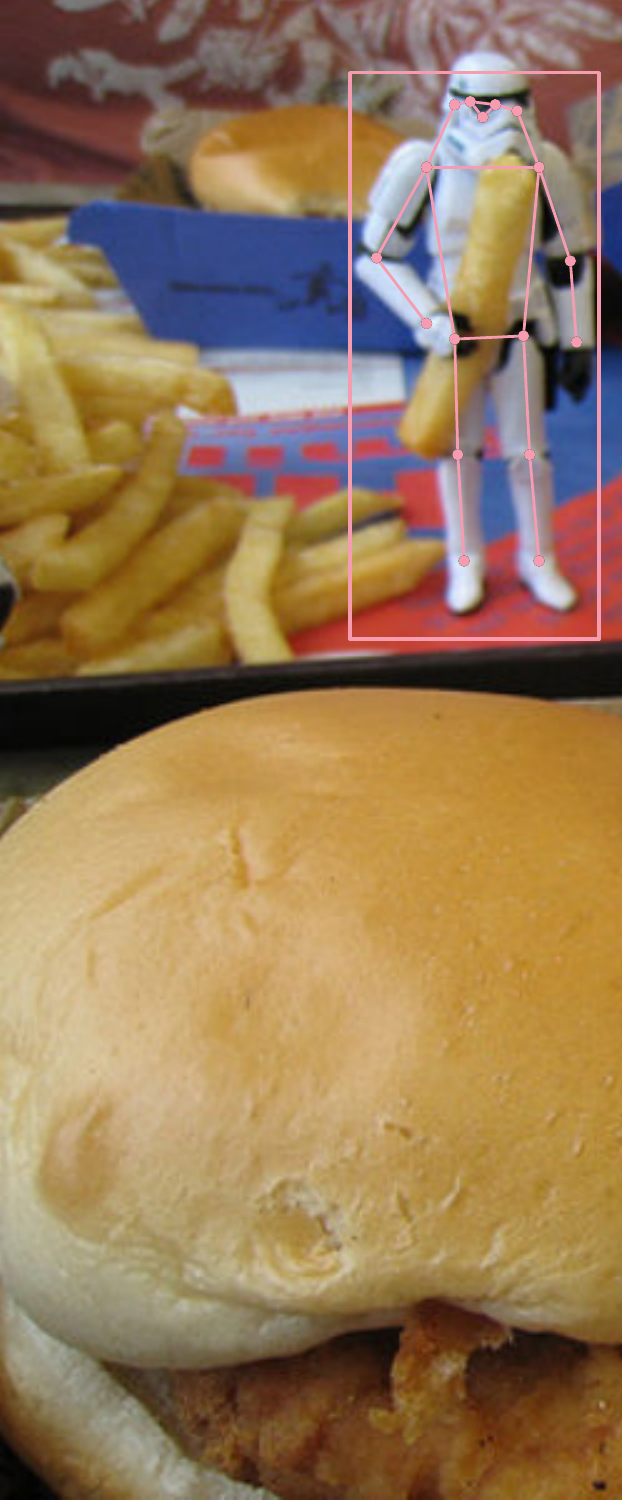}} 
  \adjustbox{height=\myheightD}
      {\includegraphics[trim={7.5cm 0 2.5cm 0},clip]{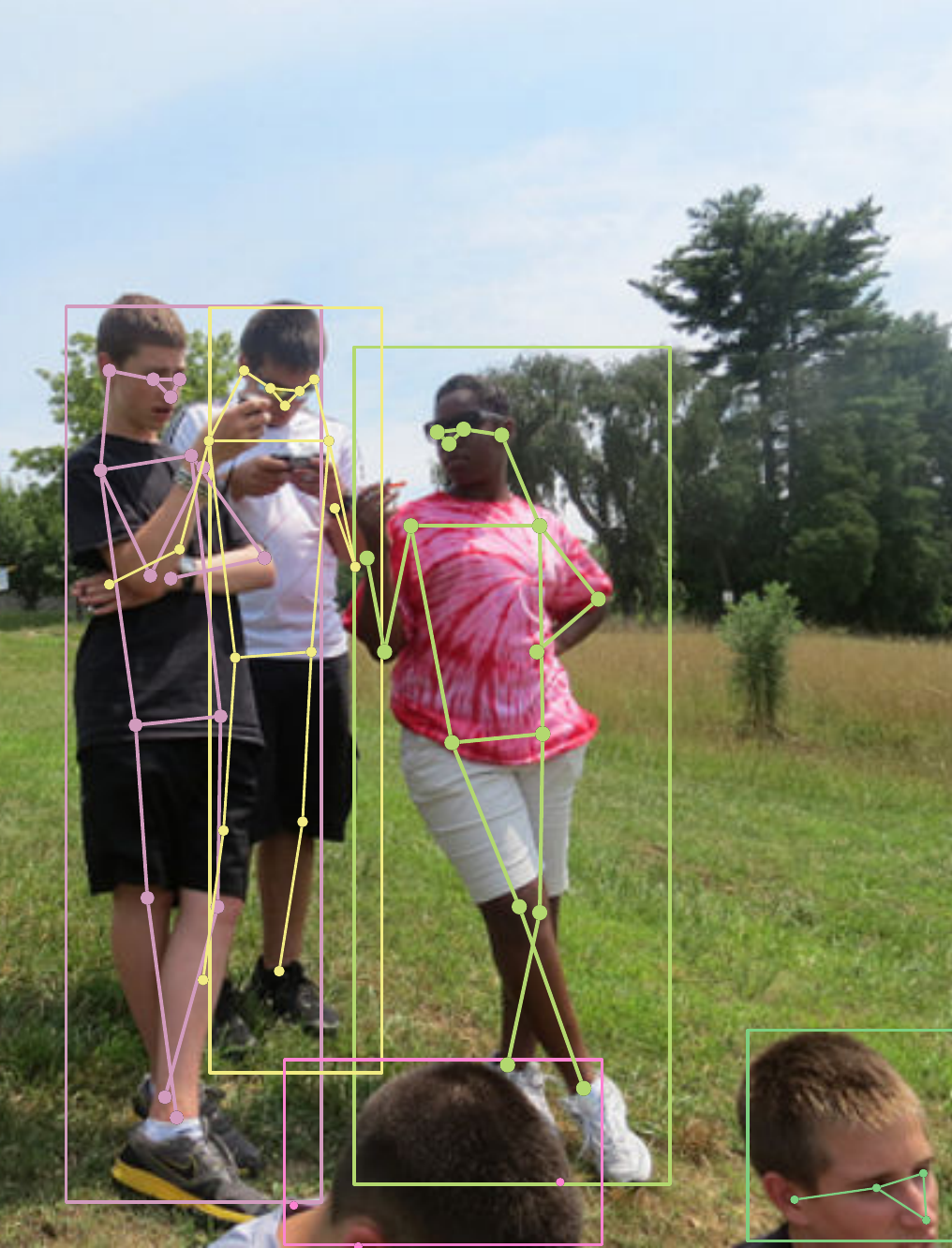}} 
  \adjustbox{height=\myheightD}
      {\includegraphics[trim={0cm 0 0cm 0},clip]{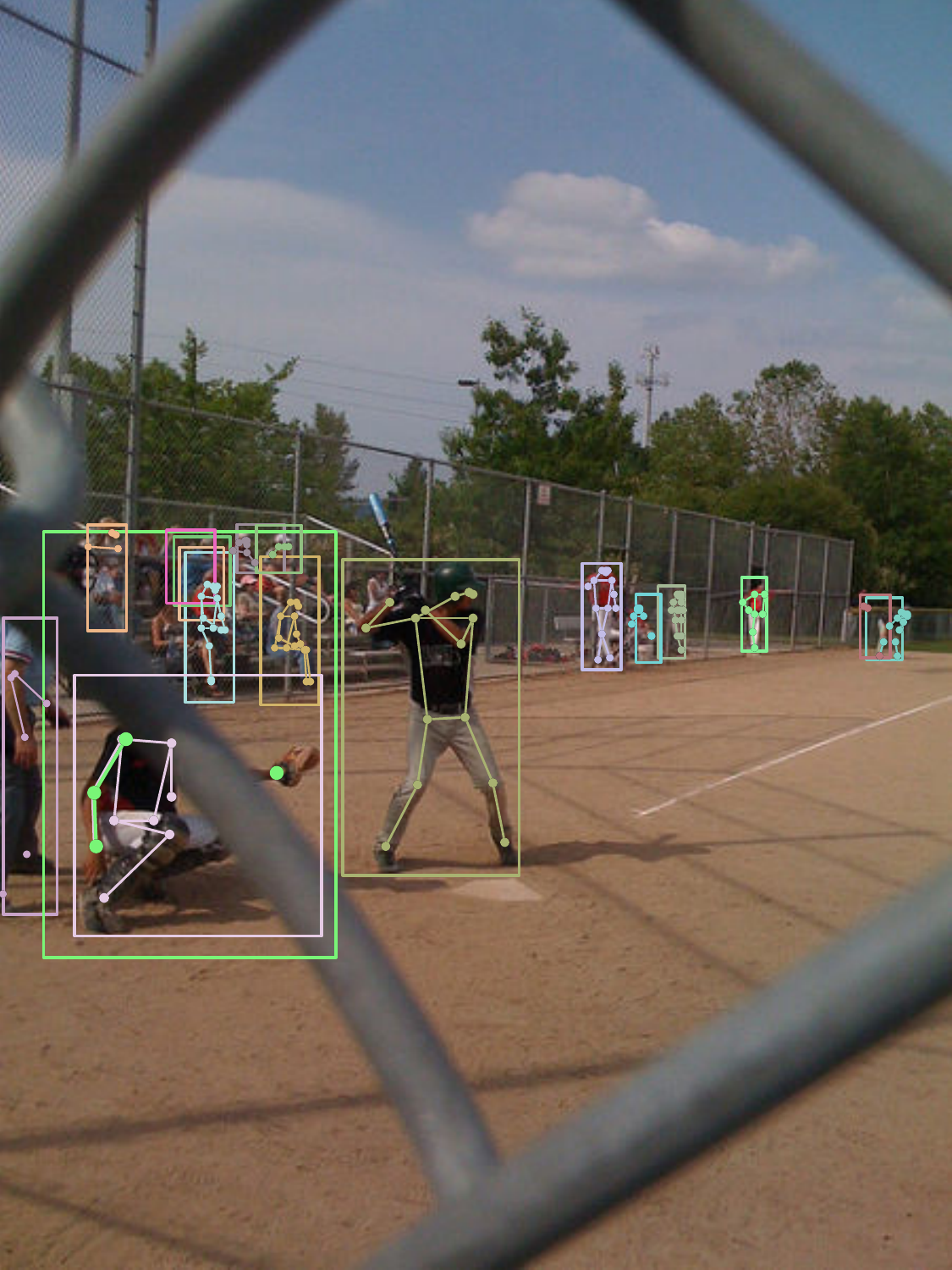}} 
  \adjustbox{height=\myheightD}
      {\includegraphics[trim={0cm 0 0cm 0},clip]{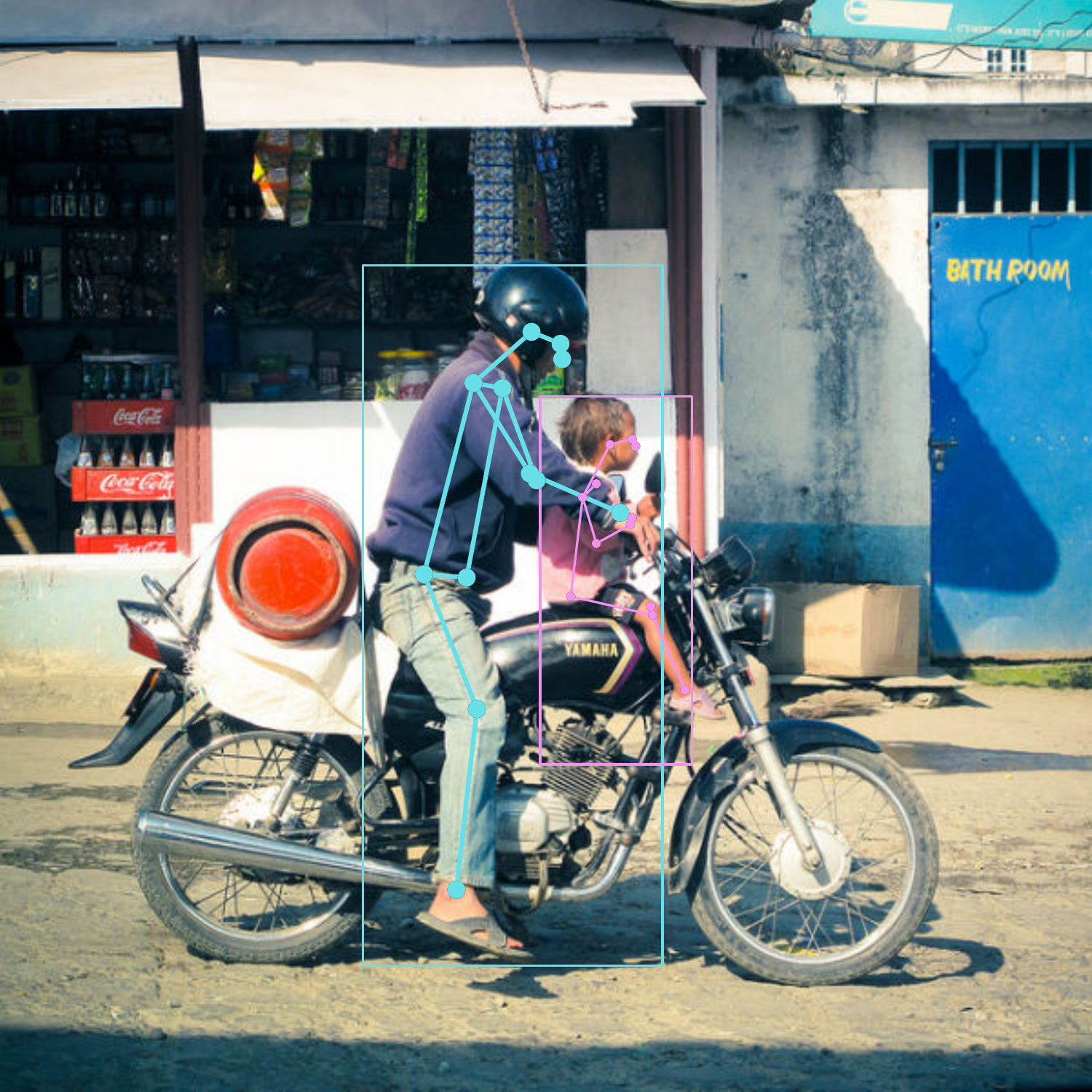}} 
      
  \vspace{1mm}
  \centering
  \adjustbox{height=\myheightE}
      {\includegraphics[trim={2cm 0 0cm 0},clip]{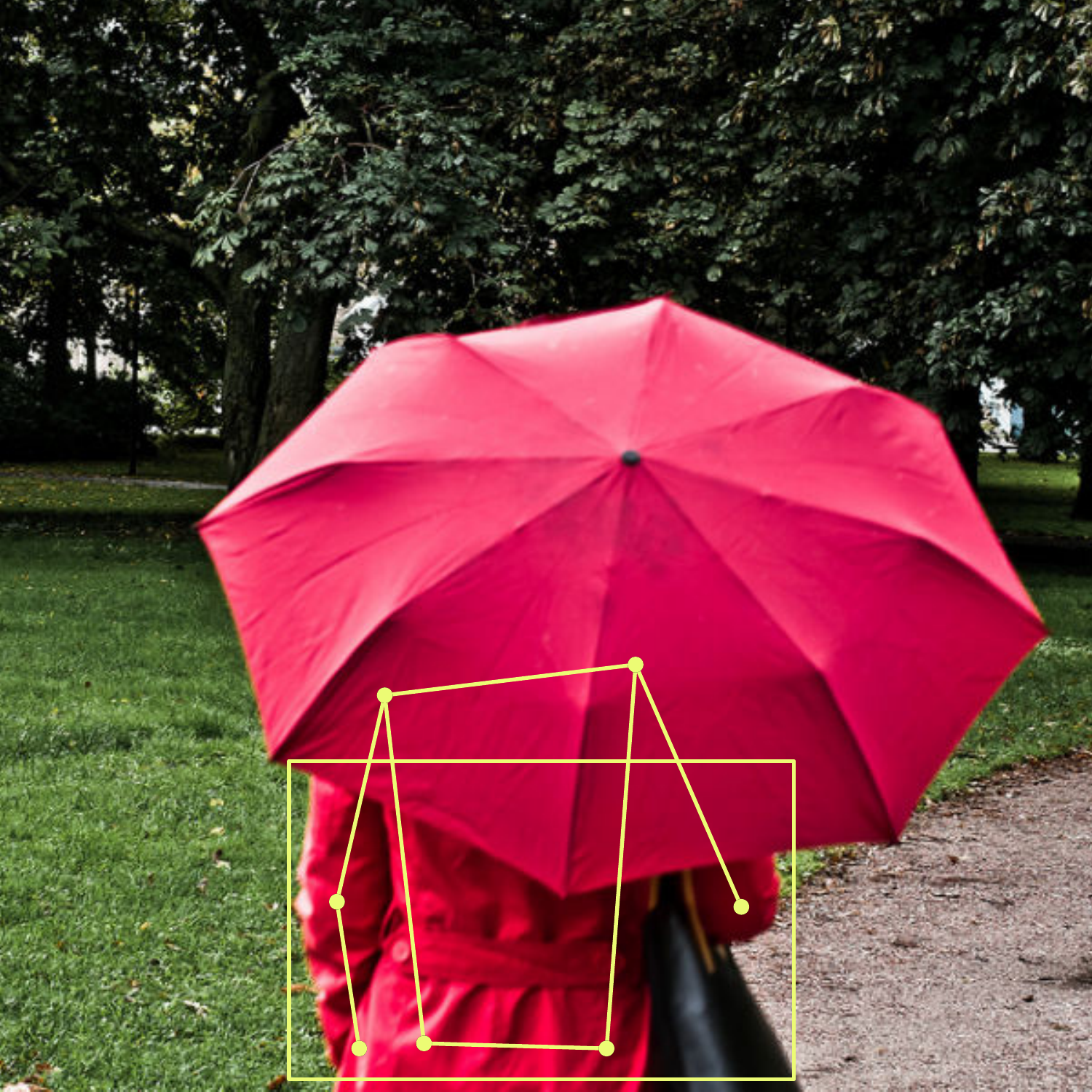}}
  \adjustbox{height=\myheightE}
      {\includegraphics[trim={0cm 0 0cm 0},clip]{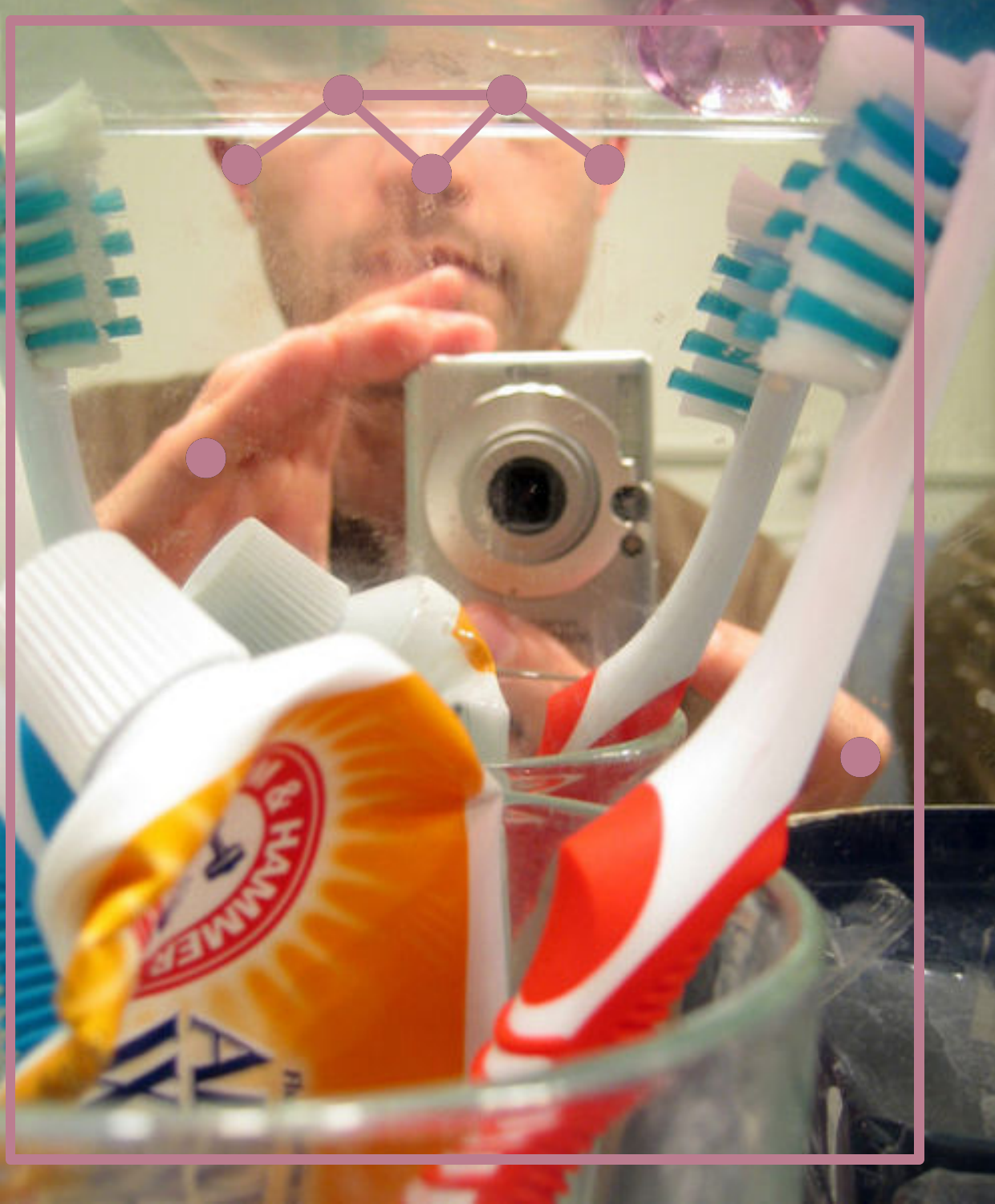}} 
  \adjustbox{height=\myheightE}
      {\includegraphics[trim={0cm 0 0cm 0},clip]{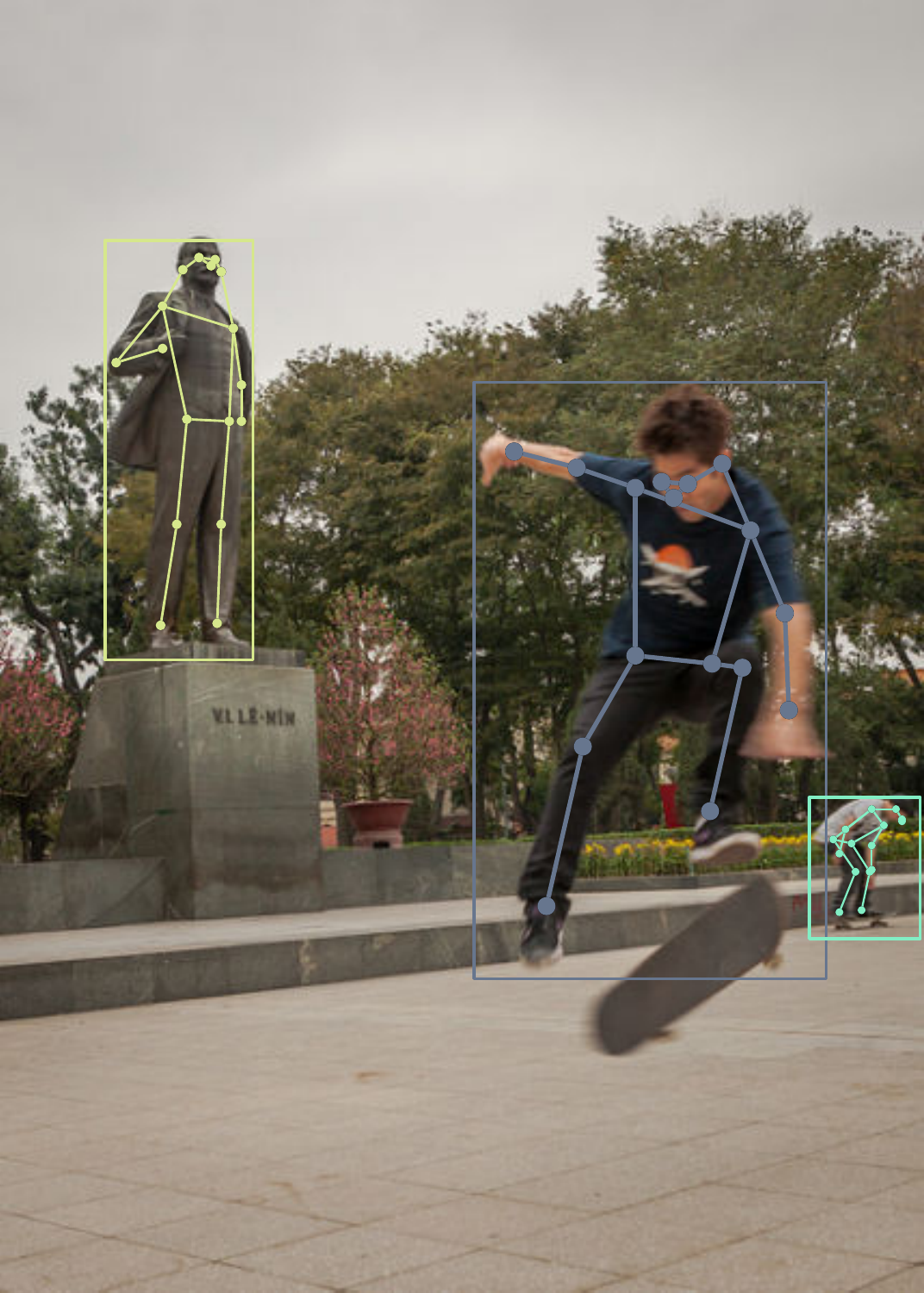}} 
  \adjustbox{height=\myheightE}
      {\includegraphics[trim={0cm 0 0cm 0},clip]{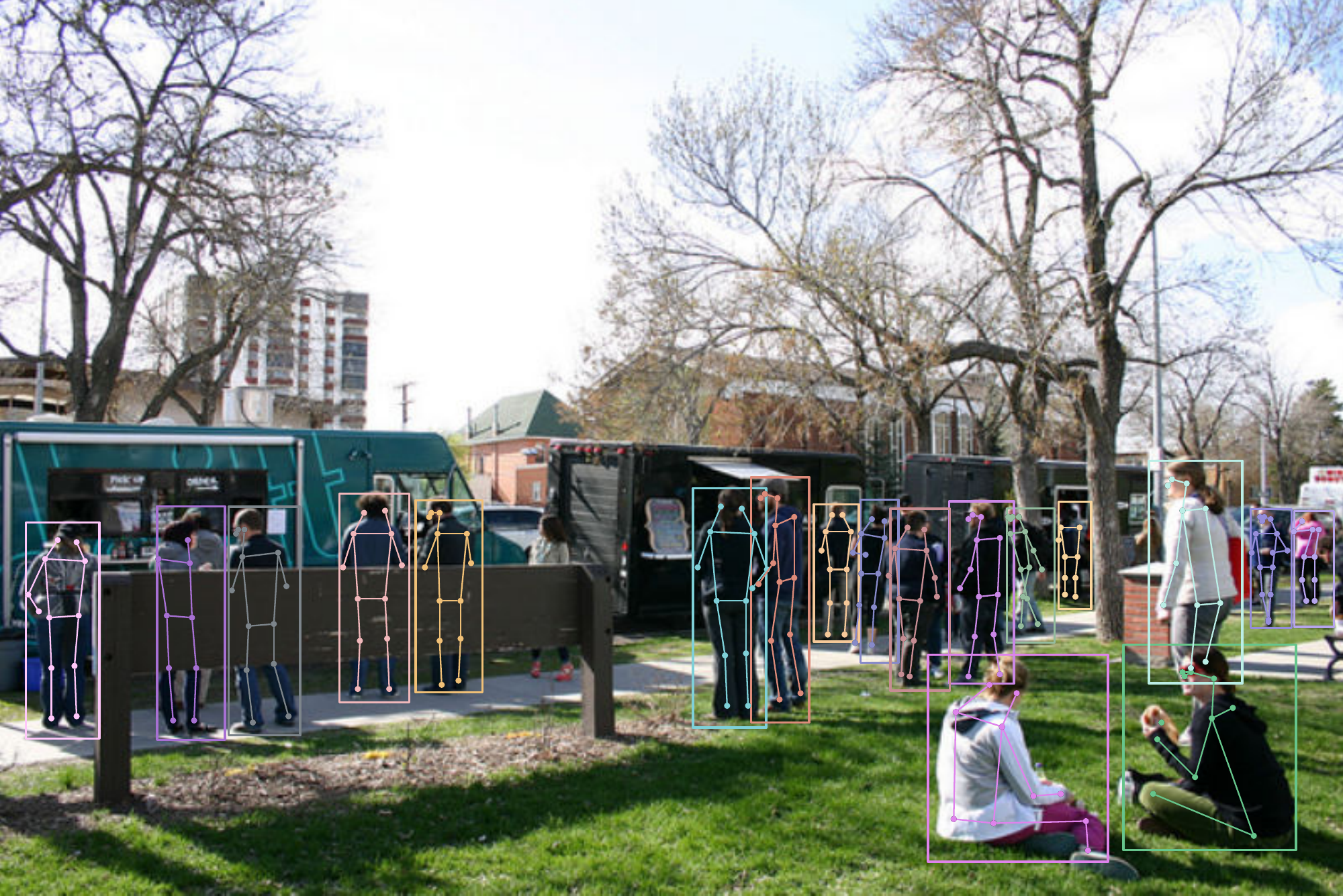}} 
  \adjustbox{height=\myheightE}
      {\includegraphics[trim={12cm 0 0cm 0},clip]{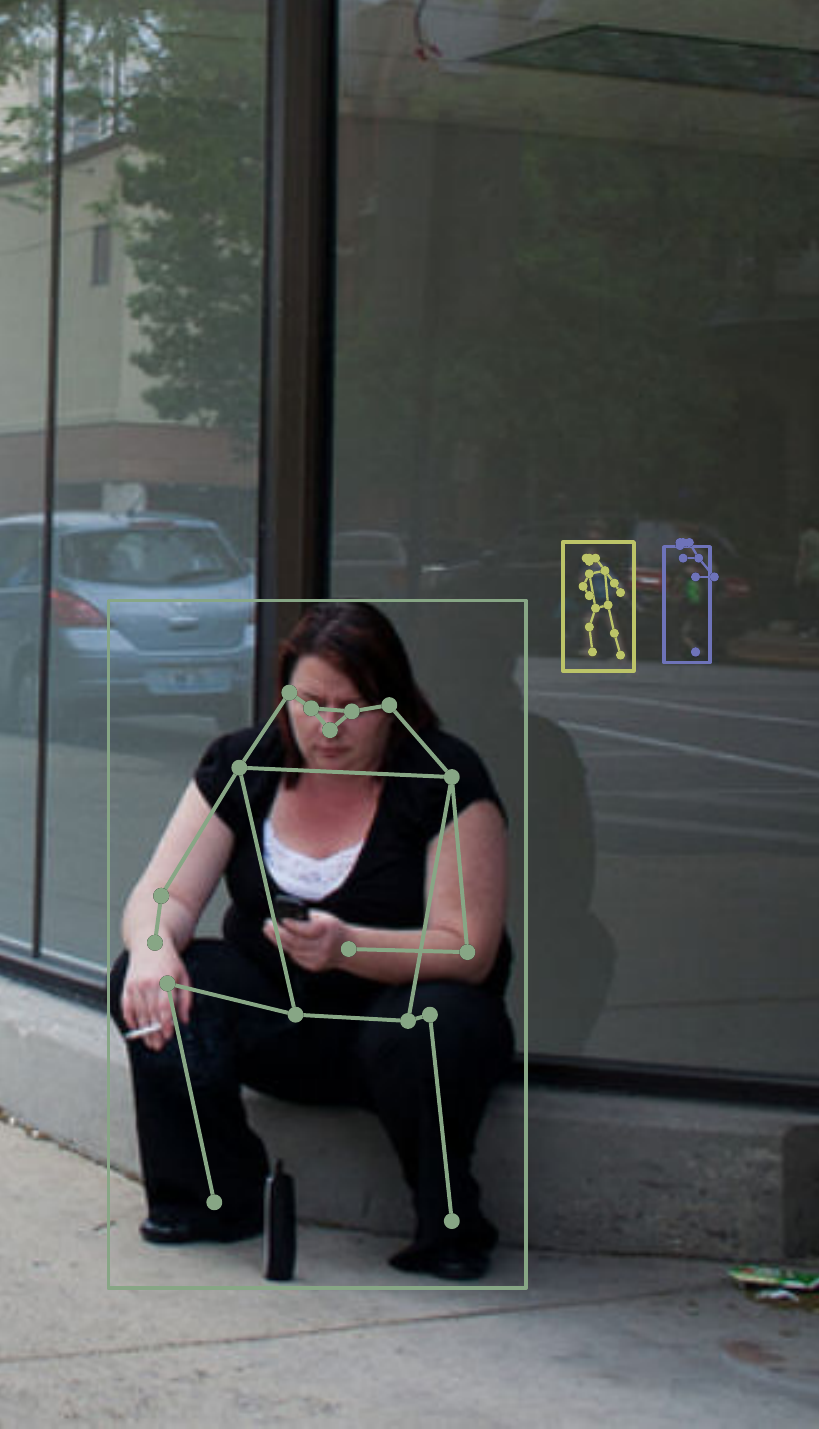}}          
      
      
\end{center}
  \caption{Detection and pose estimation results using our system on a random selection from the COCO test-dev set. For each detected person, we display the detected bounding box together with the estimated keypoints. All detections for one person are colored the same way. It is worth noting that our system works in heavily cluttered scenes (third row, rightmost and last row, right); it deals well with occlusions (last row, left) and hallucinates occluded joints. Last but not least, some of the false positive detections are in reality correct as they represent pictures of people (first row, middle) or toys (fourth row, middle). Figure best viewed zoomed in on a monitor.}
  \label{fig:results_array}
\end{figure*}

\begin{table*}
\centering
\caption{Performance on COCO keypoint \textbf{test-dev} split.}
\label{table:coco_results_testdev}
\scalebox{0.8}{
\begin{tabular}{lcccccccccc}
                  & AP    & AP .5 & AP .75 & AP (M) & AP (L) & AR    & AR .5 & AR .75 & AR (M) & AR (L) \\ \hline
CMU-Pose~\cite{cmu_mscoco}  & 0.618 & 0.849 & 0.675 & 0.571 & 0.682 & 0.665 & 0.872 & 0.718 & 0.606 & 0.746 \\
Mask-RCNN~\cite{he2017mask} & 0.631 & 0.873 & 0.687 & 0.578 & 0.714 &  &  &  &  &  \\ \hline
G-RMI (ours): \emph{COCO-only} & 0.649 & 0.855 & 0.713 & 0.623 & 0.700 & 0.697 & 0.887 & 0.755 & 0.644 & 0.771 \\
G-RMI (ours): \emph{COCO+int}  & \textbf{0.685} & 0.871 & 0.755  & 0.658  & 0.733  & 0.733 & 0.901 & 0.795  & 0.681  & 0.804 \\
\end{tabular}}
\end{table*}

\begin{table*}
\centering
\caption{Performance on COCO keypoint \textbf{test-standard} split.}
\label{table:coco_results_teststandard}
\scalebox{0.8}{
\begin{tabular}{lcccccccccc}
                  & AP    & AP .5 & AP .75 & AP (M) & AP (L) & AR    & AR .5 & AR .75 & AR (M) & AR (L) \\ \hline
CMU-Pose\cite{cmu_mscoco}  & 0.611 & 0.844 & 0.667 & 0.558 & 0.684 & 0.665 & 0.872 & 0.718 & 0.602 & 0.749 \\ \hline
G-RMI (ours): \emph{COCO-only} & 0.643 & 0.846 & 0.704 & 0.614 & 0.696 & 0.698 & 0.885 & 0.755 & 0.644 & 0.771 \\
G-RMI (ours): \emph{COCO+int}  & \textbf{0.673} & 0.854 & 0.735 & 0.642 & 0.726 & 0.730 & 0.898 & 0.789 & 0.675 & 0.805 \\
\end{tabular}}
\end{table*}

\subsection{Experimental Setup}
\label{sec:experimental_setup}

We have implemented out system in Tensorflow \cite{tensorflow2015-whitepaper}. We use distributed training across several machines equipped with Tesla K40 GPUs.

For person detector training we use 9 GPUs. We optimize with asynchronous SGD with momentum set to $0.9$. The learning rate starts at $0.0003$ and is decreased by a factor of 10 at $800$k steps. We train for 1M steps.

For pose estimator training we use two machines equipped with 8 GPUs each and batch size equal to 24 (3 crops per GPU times 8 GPUs). We use a fixed learning rate of $0.005$ and Polyak-Ruppert parameter averaging, which amounts to using during evaluation a running average of the parameters during training. We train for $800$k steps.

All our networks are pre-trained on the Imagenet classification dataset~\cite{imagenet2015}. To train our system we use two dataset variants; one that uses only COCO data (\emph{COCO-only}), and one that appends to this dataset samples from an internal dataset (\emph{COCO+int}). For the \emph{COCO-only} dataset we use the COCO keypoint annotations~\cite{lin2014microsoft}: From the 66,808 images (273,469 person instances) in the COCO \emph{train+val} splits, we use 62,174 images (105,698 person instances) in \emph{COCO-only} model training and use the remaining 4,301 annotated images as \emph{mini-val} evaluation set. Our \emph{COCO+int} training set is the union of \emph{COCO-only} with an additional 73,024 images randomly selected from Flickr. This in-house dataset contains an additional 227,029 person instances annotated with keypoints following a procedure similar to that described by Lin et al.~\cite{keypointchallenge}. The additional training images have been verified to have no overlap with the COCO training, validation or test sets.

We have trained our Faster-RCNN person box detection module exclusively on the \emph{COCO-only} dataset. We have experimented training our ResNet-based pose estimation module either on the \emph{COCO-only} or on the augmented \emph{COCO+int} datasets and present results for both. For \emph{COCO+int} pose training we use mini-batches that contain COCO and in-house annotation instances in 1:1 ratio.

\begin{table*}[h]
\centering
\caption{Ablation on the box detection module: Performance on COCO keypoint \emph{mini-val} when using alternative box detection modules trained on \emph{COCO-only} or ground truth boxes. We use the default ResNet-101 pose estimation module trained on either \emph{COCO-only} or \emph{COCO+int}. We mark with an asterisk our default box detection module used in all other experiments.}
\label{table:detector_ablation}
\scalebox{0.8}{
\begin{tabular}{lccccccccccc}
Box Module & Poser Train & AP    & AP .5 & AP .75 & AP (M) & AP (L) & AR    & AR .5 & AR .75 & AR (M) & AR (L) \\ \hline
Faster-RCNN (600x900) & COCO-only & 0.657 & 0.831 & 0.721  & 0.617 & 0.725 & 0.699 & 0.856 & 0.754 & 0.634 & 0.788 \\
Faster-RCNN (800x1200)$^*$ & COCO-only & 0.667 & 0.851 & 0.730  & 0.633 & 0.726 & 0.708 & 0.874 & 0.763 & 0.652 & 0.786 \\
Ground-truth boxes    & COCO-only & 0.704 & 0.904 & 0.771  & 0.684 & 0.746 & 0.736 & 0.911 & 0.794 & 0.693 & 0.796 \\ \hline
Faster-RCNN (600x900)  & COCO+int & 0.693 & 0.854 & 0.757  & 0.650 & 0.762 & 0.730 & 0.871 & 0.786 & 0.665 & 0.819 \\
Faster-RCNN (800x1200)$^*$ &COCO+int & 0.700 & 0.860 & 0.764  & 0.665 & 0.760 & 0.742 & 0.888 & 0.800 & 0.686 & 0.820 \\
Ground-truth boxes     & COCO+int & 0.745 & 0.925 & 0.815  & 0.725 & 0.783 & 0.774 & 0.930 & 0.835 & 0.735 & 0.831 \\
\end{tabular}}
\end{table*}

\begin{table*}[h]
\centering
\caption{Ablation on the pose estimation module: Performance on COCO keypoint \emph{test-dev} when using alternative pose estimation modules trained on \emph{COCO+int}. We use the default ResNet-101 box detection module trained on \emph{COCO-only}. We mark with an asterisk our default pose estimation module used in all other experiments.}
\label{table:poser_ablation}
\scalebox{0.8}{
\begin{tabular}{lccccccccccc}
Pose Module & Poser Train & AP    & AP .5 & AP .75 & AP (M) & AP (L) & AR    & AR .5 & AR .75 & AR (M) & AR (L) \\ \hline
ResNet-50 (257x185) & COCO+int   & 0.649 & 0.853 & 0.722  & 0.627 & 0.693 & 0.699 & 0.890 & 0.763 & 0.650 & 0.766 \\
ResNet-50 (353x257) & COCO+int   & 0.666 & 0.862 & 0.734  & 0.638 & 0.717 & 0.714 & 0.894 & 0.774 & 0.661 & 0.787 \\
ResNet-101 (257x185)& COCO+int   & 0.661 & 0.862 & 0.734 & 0.641 & 0.708 & 0.712 & 0.895 & 0.777 & 0.662 & 0.782 \\
ResNet-101 (353x257)$^*$& COCO+int & 0.685 & 0.871 & 0.755 & 0.658 & 0.733 & 0.733 & 0.901 & 0.795 & 0.681 & 0.804 \\
\end{tabular}}
\end{table*}

\subsection{COCO Keypoints Detection State-of-the-Art}
\label{sec:coco_state_of_the_art}

Table~\ref{table:coco_results_testdev} shows the COCO keypoint test-dev split performance of our system trained on \emph{COCO-only} or trained on \emph{COCO+int} datasets. A random selection of test-dev inference samples are shown in Figure~\ref{fig:results_array}.

Table~\ref{table:coco_results_teststandard} shows the COCO keypoint test-standard split results of our model with the pose estimator trained on either \emph{COCO-only} or \emph{COCO+int} training set.

Even with \emph{COCO-only} training, we achieve state-of-the-art results on the COCO test-dev and test-standard splits, outperforming the COCO 2016 challenge winning CMU-Pose team \cite{cmu_mscoco} and the very recent Mask-RCNN method \cite{he2017mask}. Our best results are achieved with the pose estimator trained on \emph{COCO+int} data, yielding an AP score of 0.673 on \emph{test-standard}, an absolute 6.2\% improvement over the 0.611 \emph{test-standard} score of CMU-Pose \cite{cmu_mscoco}.

\subsection{Ablation Study: Box Detection Module}

An important question for our two-stage system is its sensitivity to the quality of its box detection and pose estimator constituent modules. We examine two variants of the ResNet-101 based Faster-RCNN person box detector, (a) a fast 600x900 variant that uses input images with small side 600 pixels and large side 900 pixels and (b) an accurate 800x1200 variant that uses input images with small side 800 pixels and large side 1200 pixels. Their box detection AP on our COCO person \emph{mini-val} is 0.466 and 0.500, respectively. Their box detection AP on COCO \emph{test-dev} is 0.456 and 0.487, respectively. For reference, the person box detection AP on COCO \emph{test-dev} of the top-performing multi-crop/ensemble entry of \cite{huang2016speed} is 0.539. We have also tried feeding our pose estimator module with the ground truth person boxes to examine its oracle performance limit in isolation from the box detection module. We report our COCO \emph{mini-val} results in Table~\ref{table:detector_ablation} for pose estimators trained on either \emph{COCO-only} or \emph{COCO+int}. We use the accurate Faster-RCNN (800x1200) box detector for all results in the rest of the paper.

\subsection{Ablation Study: Pose Estimation Module}

We have experimented with alternative CNN setups for our pose estimation module. We have explored CNN network backbones based on either the faster ResNet-50 or the more accurate ResNet-101, while keeping ResNet-101 as CNN backbone for the Faster-RCNN box detection module. We have also experimented with two sizes for the image crops that are fed as input to the pose estimator: Smaller (257x185) for faster inference or larger (353x257) for higher accuracy. We report in Table~\ref{table:poser_ablation} COCO \emph{test-dev} results for the four CNN backbone/ crop size combinations, using \emph{COCO+int} for pose estimator training. We see that ResNet-101 performs about 2\% better but in computation-constrained environments ResNet-50 remains a competitive alternative. We use the accurate ResNet-101 (353x257) pose estimator with disk radius $R=25$ pixels in the rest of the paper.

\subsection{OKS-Based Non Maximum Suppression}

We examine the effect of the proposed OKS-based non-maximum suppression method at the output of the pose estimator for different values of the OKS-NMS threshold. In all experiments the value of the IOU-NMS threshold at the output of the person box detector remains fixed at 0.6. We report in Table~\ref{table:oks_nms} COCO \emph{mini-val} results using either \emph{COCO-only} or \emph{COCO+int} for pose estimator training. We fix the OKS-NMS threshold to 0.5 in the rest of the paper.

\begin{table}
\centering
\caption{Performance (AP) on COCO keypoint \emph{mini-val} with varying values for the OKS-NMS threshold. The pose estimator has been trained with either \emph{COCO-only} or \emph{COCO+int} data.}
\label{table:oks_nms}
\scalebox{0.8}{
\begin{tabular}{lccccc}
Threshold             & 0.1 & 0.3 & 0.5$^*$ & 0.7 & 0.9 \\ \hline
AP (\emph{COCO-only}) & 0.638 & 0.664 & 0.667 & 0.665 & 0.658 \\
AP (\emph{COCO+int})  & 0.672 & 0.699 & 0.700 & 0.701 & 0.694 \\
\end{tabular}}
\end{table}

\section{Conclusion}

In this work we address the problem of person detection and pose estimation in cluttered images `in the wild'. We present a simple two stage system, consisting of a person detection stage followed by a keypoint estimation stage for each person. Despite its simplicity it achieves state-of-art results as measured on the challenging COCO benchmark.

\eat{
In the future, we plan to investigate  systems which share features between the detector and keypoint estimation and can be trained end-to-end. In this way we could hopefully boost our performance further while reducing computation at inference time.
}

\subsection*{Acknowledgments}

We are grateful to the authors of \cite{huang2016speed} for making their excellent Faster-RCNN implementation available to us. We would like to thank Hartwig Adam for encouraging and supporting this project and Akshay Gogia and Gursheesh Kour for managing our internal annotation effort.

{\small
\bibliographystyle{ieee}
\bibliography{egbib}

\begin{thebibliography}{10}\itemsep=-1pt

\bibitem{tensorflow2015-whitepaper}
M.~Abadi, A.~Agarwal, P.~Barham, E.~Brevdo, et~al.
\newblock {TensorFlow}: Large-scale machine learning on heterogeneous systems,
  2015.
\newblock Software available from tensorflow.org.

\bibitem{andriluka14cvpr}
M.~Andriluka, L.~Pishchulin, P.~Gehler, and B.~Schiele.
\newblock 2d human pose estimation: New benchmark and state of the art
  analysis.
\newblock In {\em CVPR}, 2014.

\bibitem{andriluka2009pictorial}
M.~Andriluka, S.~Roth, and B.~Schiele.
\newblock {Pictorial structures revisited: People detection and articulated
  pose estimation}.
\newblock In {\em CVPR}, 2009.

\bibitem{belagiannis20143d}
V.~Belagiannis, S.~Amin, M.~Andriluka, B.~Schiele, N.~Navab, and S.~Ilic.
\newblock 3d pictorial structures for multiple human pose estimation.
\newblock In {\em CVPR}, pages 1669--1676, 2014.

\bibitem{belagiannis20153d}
V.~Belagiannis, S.~Amin, M.~Andriluka, B.~Schiele, N.~Navab, and S.~Ilic.
\newblock 3d pictorial structures revisited: Multiple human pose estimation.
\newblock In {\em CVPR}, 2015.

\bibitem{zisserman2016}
V.~Belagiannis and A.~Zisserman.
\newblock Recurrent human pose estimation.
\newblock In {\em arxiv}, 2016.

\bibitem{bulat2016}
A.~Bulat and G.~Tzimiropoulos.
\newblock Human pose estimation via convolutional part heatmap regression.
\newblock In {\em ECCV}, 2016.

\bibitem{cmu_mscoco}
Z.~Cao, T.~Simon, S.-E. Wei, and Y.~Sheikh.
\newblock Realtime multi-person 2d pose estimation using part affinity fields.
\newblock {\em arXiv:1611.08050v1}, 2016.

\bibitem{chen2016deeplab}
L.-C. Chen, G.~Papandreou, I.~Kokkinos, K.~Murphy, and A.~L. Yuille.
\newblock Deeplab: Semantic image segmentation with deep convolutional nets,
  atrous convolution, and fully connected crfs.
\newblock {\em arXiv:1606.00915}, 2016.

\bibitem{Chen_NIPS14}
X.~Chen and A.~Yuille.
\newblock Articulated pose estimation by a graphical model with image dependent
  pairwise relations.
\newblock In {\em NIPS}, 2014.

\bibitem{dantone13cvpr}
M.~Dantone, J.~Gall, C.~Leistner, and L.~V. Gool.
\newblock Human pose estimation using body parts dependent joint regressors.
\newblock In {\em CVPR}, 2013.

\bibitem{BetterAppearancePic}
M.~Eichner and V.~Ferrari.
\newblock Better appearance models for pictorial structures.
\newblock In {\em BMVC}, 2009.

\bibitem{eichner2010we}
M.~Eichner and V.~Ferrari.
\newblock We are family: Joint pose estimation of multiple persons.
\newblock In {\em ECCV}, pages 228--242. Springer, 2010.

\bibitem{erhan2014scalable}
D.~Erhan, C.~Szegedy, A.~Toshev, and D.~Anguelov.
\newblock Scalable object detection using deep neural networks.
\newblock In {\em CVPR}, pages 2147--2154, 2014.

\bibitem{FelzenszwalbDPM}
P.~Felzenszwalb, D.~McAllester, and D.~Ramanan.
\newblock A discriminatively trained, multiscale, deformable part model.
\newblock In {\em CVPR}, 2008.

\bibitem{Fischler73}
M.~A. Fischler and R.~Elschlager.
\newblock The representation and matching of pictorial structures.
\newblock In {\em IEEE TOC}, 1973.

\bibitem{girshick2014rich}
R.~Girshick, J.~Donahue, T.~Darrell, and J.~Malik.
\newblock Rich feature hierarchies for accurate object detection and semantic
  segmentation.
\newblock In {\em CVPR}, pages 580--587, 2014.

\bibitem{armlets2013}
G.~Gkioxari, P.~Arbelaez, L.~Bourdev, and J.~Malik.
\newblock Articulated pose estimation using discriminative armlet classifiers.
\newblock In {\em CVPR}, 2013.

\bibitem{gkioxari2014using}
G.~Gkioxari, B.~Hariharan, R.~Girshick, and J.~Malik.
\newblock Using k-poselets for detecting people and localizing their keypoints.
\newblock In {\em CVPR}, pages 3582--3589, 2014.

\bibitem{chain16}
G.~Gkioxari, A.~Toshev, and N.~Jaitly.
\newblock Chained predictions using convolutional neural networks.
\newblock In {\em ECCV}, 2016.

\bibitem{he2017mask}
K.~He, G.~Gkioxari, P.~Doll{\'a}r, and R.~Girshick.
\newblock Mask r-cnn.
\newblock {\em arXiv:1703.06870v2}, 2017.

\bibitem{He2016ResNets}
K.~He, X.~Zhang, S.~Ren, and J.~Sun.
\newblock Deep residual learning for image recognition.
\newblock In {\em {CVPR}}, 2016.

\bibitem{huang2016speed}
J.~Huang, V.~Rathod, C.~Sun, M.~Zhu, A.~Korattikara, A.~Fathi, I.~Fischer,
  Z.~Wojna, Y.~Song, S.~Guadarrama, et~al.
\newblock Speed/accuracy trade-offs for modern convolutional object detectors.
\newblock {\em arXiv:1611.10012}, 2016.

\bibitem{insafutdinov2016articulated}
E.~Insafutdinov, M.~Andriluka, L.~Pishchulin, S.~Tang, B.~Andres, and
  B.~Schiele.
\newblock Articulated multi-person tracking in the wild.
\newblock {\em arXiv:1612.01465}, 2016.

\bibitem{deeper_cut}
E.~Insafutdinov, L.~Pishchulin, B.~Andres, M.~Andriluka, and B.~Schiele.
\newblock Deepercut: A deeper, stronger, and faster multi-person pose
  estimation model.
\newblock In {\em ECCV}, 2016.

\bibitem{iqbal2016multi}
U.~Iqbal and J.~Gall.
\newblock Multi-person pose estimation with local joint-to-person associations.
\newblock In {\em ECCV}, pages 627--642. Springer, 2016.

\bibitem{jainiclr2014}
A.~Jain, J.~Tompson, M.~Andriluka, G.~Taylor, and C.~Bregler.
\newblock Learning human pose estimation features with convolutional networks.
\newblock In {\em ICLR}, 2014.

\bibitem{johnson11cvpr}
S.~Johnson and M.~Everingham.
\newblock {Learning Effective Human Pose Estimation from Inaccurate
  Annotation}.
\newblock In {\em CVPR}, 2011.

\bibitem{ladicky2013human}
L.~Ladicky, P.~H. Torr, and A.~Zisserman.
\newblock Human pose estimation using a joint pixel-wise and part-wise
  formulation.
\newblock In {\em CVPR}, pages 3578--3585, 2013.

\bibitem{li2016r}
Y.~Li, K.~He, J.~Sun, et~al.
\newblock {R-FCN}: Object detection via region-based fully convolutional
  networks.
\newblock In {\em Advances in Neural Information Processing Systems}, pages
  379--387, 2016.

\bibitem{keypointchallenge}
T.-Y. Lin, Y.~Cui, G.~Patterson, M.~R. Ronchi, L.~Bourdev, R.~Girshick, and
  P.~Doll{\'a}r.
\newblock Coco 2016 keypoint challenge.
\newblock 2016.

\bibitem{lin2014microsoft}
T.-Y. Lin, M.~Maire, S.~Belongie, J.~Hays, P.~Perona, D.~Ramanan,
  P.~Doll{\'a}r, and C.~L. Zitnick.
\newblock Microsoft coco: Common objects in context.
\newblock In {\em ECCV}, pages 740--755. Springer, 2014.

\bibitem{stackedhourglass}
A.~Newell, K.~Yang, and J.~Deng.
\newblock Stacked hourglass networks for human pose estimation.
\newblock In {\em ECCV}, 2016.

\bibitem{pishchulin13cvpr}
L.~Pishchulin, M.~Andriluka, P.~Gehler, and B.~Schiele.
\newblock Poselet conditioned pictorial structures.
\newblock In {\em CVPR}, 2013.

\bibitem{deepcut}
L.~Pishchulin, E.~Insafutdinov, S.~Tang, B.~Andres, M.~Andriluka, P.~Gehler,
  and B.~Schiele.
\newblock Deepcut: Joint subset partition and labeling for multi person pose
  estimation.
\newblock In {\em CVPR}, 2016.

\bibitem{pishchulin2012articulated}
L.~Pishchulin, A.~Jain, M.~Andriluka, T.~Thorm{\"a}hlen, and B.~Schiele.
\newblock Articulated people detection and pose estimation: Reshaping the
  future.
\newblock In {\em CVPR}, pages 3178--3185, 2012.

\bibitem{Ren2015}
S.~Ren, K.~He, R.~Girshick, and J.~Sun.
\newblock Faster {R-CNN}: Towards {Real-Time} object detection with region
  proposal networks.
\newblock In {\em {NIPS}}, 2015.

\bibitem{imagenet2015}
O.~Russakovsky, J.~Deng, H.~Su, J.~Krause, S.~Satheesh, S.~Ma, Z.~Huang,
  A.~Karpathy, A.~Khosla, M.~Bernstein, A.~C. Berg, and L.~Fei-Fei.
\newblock {ImageNet Large Scale Visual Recognition Challenge}.
\newblock {\em IJCV}, 115(3):211--252, 2015.

\bibitem{Sapp2010}
B.~Sapp, C.~Jordan, and B.Taskar.
\newblock Adaptive pose priors for pictorial structures.
\newblock In {\em CVPR}, 2010.

\bibitem{modec2013}
B.~Sapp and B.~Taskar.
\newblock Modec: Multimodal decomposable models for human pose estimation.
\newblock In {\em CVPR}, 2013.

\bibitem{sun2011articulated}
M.~Sun and S.~Savarese.
\newblock Articulated part-based model for joint object detection and pose
  estimation.
\newblock In {\em ICCV}, pages 723--730, 2011.

\bibitem{szegedy2016inception}
C.~Szegedy, S.~Ioffe, and V.~Vanhoucke.
\newblock Inception-v4, inception-resnet and the impact of residual connections
  on learning.
\newblock {\em arXiv:1602.07261}, 2016.

\bibitem{szegedy2015going}
C.~Szegedy, W.~Liu, Y.~Jia, P.~Sermanet, S.~Reed, D.~Anguelov, D.~Erhan,
  V.~Vanhoucke, and A.~Rabinovich.
\newblock Going deeper with convolutions.
\newblock In {\em CVPR}, pages 1--9, 2015.

\bibitem{tompsonnips2014}
J.~Tompson, A.~Jain, Y.~LeCun, and C.~Bregler.
\newblock Join training of a convolutional network and a graphical model for
  human pose estimation.
\newblock In {\em NIPS}, 2014.

\bibitem{deeppose}
A.~Toshev and C.~Szegedy.
\newblock Deeppose: Human pose estimation via deep neural networks.
\newblock In {\em CVPR}, 2014.

\bibitem{wei2016convolutional}
S.-E. Wei, V.~Ramakrishna, T.~Kanade, and Y.~Sheikh.
\newblock Convolutional pose machines.
\newblock In {\em arXiv}, 2016.

\bibitem{yang11cvpr}
Y.~Yang and D.~Ramanan.
\newblock Articulated pose estimation with flexible mixtures of parts.
\newblock In {\em CVPR}, 2011.

\end{thebibliography}
}

\newpage


\end{document}